%% file: locusrl_iclr2027.tex
\documentclass{article}
\usepackage{locusrl_preprint,times}

\usepackage{amsmath}
\usepackage{amssymb}
\usepackage{array}
\usepackage{booktabs}
\usepackage{graphicx}
\usepackage{microtype}
\usepackage{tabularx}
\usepackage{url}
\usepackage{placeins}
\usepackage{flafter}
\usepackage[hidelinks]{hyperref}

\newcolumntype{Y}{>{\raggedright\arraybackslash}X}
\newcommand{\rewardprovider}{\path{c4_llm_reward_v04_glm53flash_high_offline_cache_v1}}

\title{LocusRL: Diagnosing LLM Reward and Policy Interventions in Competitive Games}

\author{Chengyu Luan\textsuperscript{*,\textdagger}\quad Bo Xin\textsuperscript{*}\quad Songyan Guo\textsuperscript{*}\\
\bfseries Yuxiang Zuo\quad Ahmed Yazdan\quad Jiahang Li\quad Yicheng Liu\\[4pt]
\normalfont Harbin Institute of Technology\\[3pt]
\normalfont\small\textsuperscript{*}Equal contribution.\quad\textsuperscript{\textdagger}Corresponding author.\\
\normalfont\small\href{mailto:luancy1208@163.com}{\texttt{luancy1208@163.com}}}
\date{}

\iclrfinalcopy

\hypersetup{
  pdftitle={LocusRL: Diagnosing LLM Reward and Policy Interventions in Competitive Games},
  pdfauthor={Chengyu Luan, Bo Xin, Songyan Guo, Yuxiang Zuo, Ahmed Yazdan, Jiahang Li, Yicheng Liu},
  pdfsubject={Diagnostic evaluation of LLM reward and policy interventions},
  pdfkeywords={reinforcement learning, large language models, reward shaping, diagnostic evaluation}
}
\begin{document}
\maketitle
\begin{abstract}
Large language models can intervene in reinforcement learning through both reward design and action selection, yet aggregate performance offers an incomplete account of what these interventions actually do. Similar returns can conceal different learning mechanisms, while plausible rewards can induce undesirable behavior. We introduce LocusRL, a diagnostic framework that connects controlled reward--policy comparisons with audits of reward judgments, signal delivery, optimization objectives, and executed actions. The framework traces performance differences to testable explanations and checks targeted corrections through executable rules and counterfactual replay.

Across two evaluation batches covering ten Connect Four training seeds, we uncover seed-dependent reversals in intervention effects and show how tracing actual updates changes their interpretation: historical Qwen training operates through reward-weighted teacher-action likelihood. A separate matched three-seed reward-direction experiment distinguishes sensitivity to a learning signal from its usefulness. With terminal rewards held fixed, a sign-reversed dense oracle yields a 2.8\% aggregate win rate, compared with 57.2\% for terminal-only training and 46.7\% for the positive dense oracle. Thus, a reward can strongly influence learning without improving performance. At the decision level, counterfactual replay verifies a winning correction to a diagnosed action error. Complementary experiments in Leduc and reward-validation studies in Goofspiel extend the analysis to imperfect-information settings, revealing how reference-label definitions and validation-data exposure affect intervention assessment.

Together, these findings show why evaluating LLM interventions requires tracing how their outputs become learning signals and actions. LocusRL turns aggregate outcomes into actionable diagnoses and verifiable corrections.
\end{abstract}

\section{Introduction}

Large language models increasingly enter reinforcement learning as reward providers, action selectors, or both. The declared role of an LLM component, however, need not be the mechanism the learner actually experiences. A reward can be semantically plausible but absent from most updates; a policy proposal can be replaced by an execution fallback; a nominally signed objective can be realized as positive reweighting. Aggregate return collapses these possibilities into one outcome and therefore cannot, by itself, identify why an intervention helped or failed.

We call the discrepancy between an intervention's nominal role and its realized training/execution path a \emph{realization gap}. This gap is a validity threat for composite LLM--RL systems: the same nominal intervention can induce different realized signals or actions, while different nominal interventions can collapse to similar updates. LocusRL diagnoses this gap across five interfaces: reward semantics, signal delivery, optimization identity, action execution, and evidence validity. Each interface is tied to an inspectable object, a quantitative check, and a follow-up decision. The goal is not to force one causal story onto an arm-level score difference, but to replace an ambiguous outcome with falsifiable mechanism claims and, where the environment permits, an executable correction.

Connect Four illustrates why this distinction matters. An initial three-seed comparison records a Hybrid deficit relative to Policy, yet all ten discordant wins repeat one paired trajectory and execution tracing finds a fallback that discards an immediate win. Seven additional training seeds then produce both positive and negative contrasts, weakening any aggregate mechanism story. A separate numerical audit finds positive loss/NLL ratios where the historical signed-advantage description predicts negative coefficients. These observations target different interfaces---replication, execution, and optimization---and none is recoverable from the original win rates alone.

We test the resulting hypotheses with complementary follow-ups: common-runtime evaluation of fixed historical checkpoints, a matched reward-direction control, and separately verified on-policy Qwen training with formal-checkpoint evaluation. Complete successor enumeration also revises the tactical reward reference, separating avoidable errors from forced next-reply losses. Leduc and Goofspiel then stress the same protocol under hidden information, equilibrium-dependent targets, and prior validation exposure.

Our contributions are threefold. \textbf{First, an operational diagnostic protocol.} LocusRL connects five realization interfaces to explicit evidence, quantitative checks, and next actions. \textbf{Second, mechanism findings that aggregate scores hide.} Controlled and seed-wise analyses separate reward influence from reward benefit, declared from realized optimization, and legal from tactically correct execution. \textbf{Third, auditable diagnosis-to-correction instances.} We provide successor-revalidated reward references, terminal-replay certificates, and exposure-aware reward-release decisions. The novelty lies in the connected diagnosis path---how evidence at one interface changes the interpretation of the result and determines the next experiment or correction---rather than in any single logging check.

\section{Related Work}

\paragraph{LLMs as rewards and policies.}
Language models provide goals, proxy rewards, and preferences \citep{du2023ellm,kwon2023rewarddesign,klissarov2024motif}; Text2Reward and Eureka synthesize executable reward programs \citep{xie2024text2reward,ma2024eureka}. GLAM, RAGEN, and SPIRAL study language-policy learning \citep{carta2023glam,ragen2025,spiral}, while RLAnything jointly adapts components \citep{rlanything}. Potential-based shaping preserves optimal policies under stated conditions \citep{ng1999shaping}; LLM variants study this construction \citep{hounwanou2026policyinvariant}. These methods motivate LLM integration. LocusRL examines how a supplied intervention is actually used by the composed learner, connecting its outputs to updates and actions.

\paragraph{Objective and reward auditing.}
The Alignment Auditor uses inferred objectives and uncertainty to guide reward refinement \citep{bou2026alignment}; hidden-objective audits investigate deliberately concealed goals \citep{marks2025auditing}; IR$^3$ reconstructs implicit rewards from response contrasts \citep{beigi2026ir3}; and RATE examines reward-model sensitivity to response attributes \citep{reber2025rate}. Spurious Rewards studies learning responses to alternative reward signals \citep{shao2025spurious}. Reward misspecification and gaming are recognized failure modes \citep{amodei2016concrete,skalse2022defining}, and seed-level variance complicates deep RL comparisons \citep{henderson2018matters,agarwal2021precipice}. Collectively, these lines analyze reward/objective informativeness, implicit objectives, or behavior under reward variation. LocusRL is complementary but shifts the unit of analysis to the \emph{realization path}: a supplied intervention is traced through delivery, the numerical update, the execution wrapper, and evaluation provenance. OpenSpiel and PettingZoo provide game interfaces \citep{lanctot2019openspiel,terry2021pettingzoo}, and NFSP provides a classical imperfect-information learning reference \citep{heinrich2016nfsp}. Appendix~\ref{app:prior-work-map} gives a component-wise comparison.

\section{Diagnostic Protocol and Experimental Design}
\label{sec:protocol}

\subsection{Comparison and Realization Contract}
\begin{figure}[!t]
\centering
\includegraphics[width=0.90\linewidth]{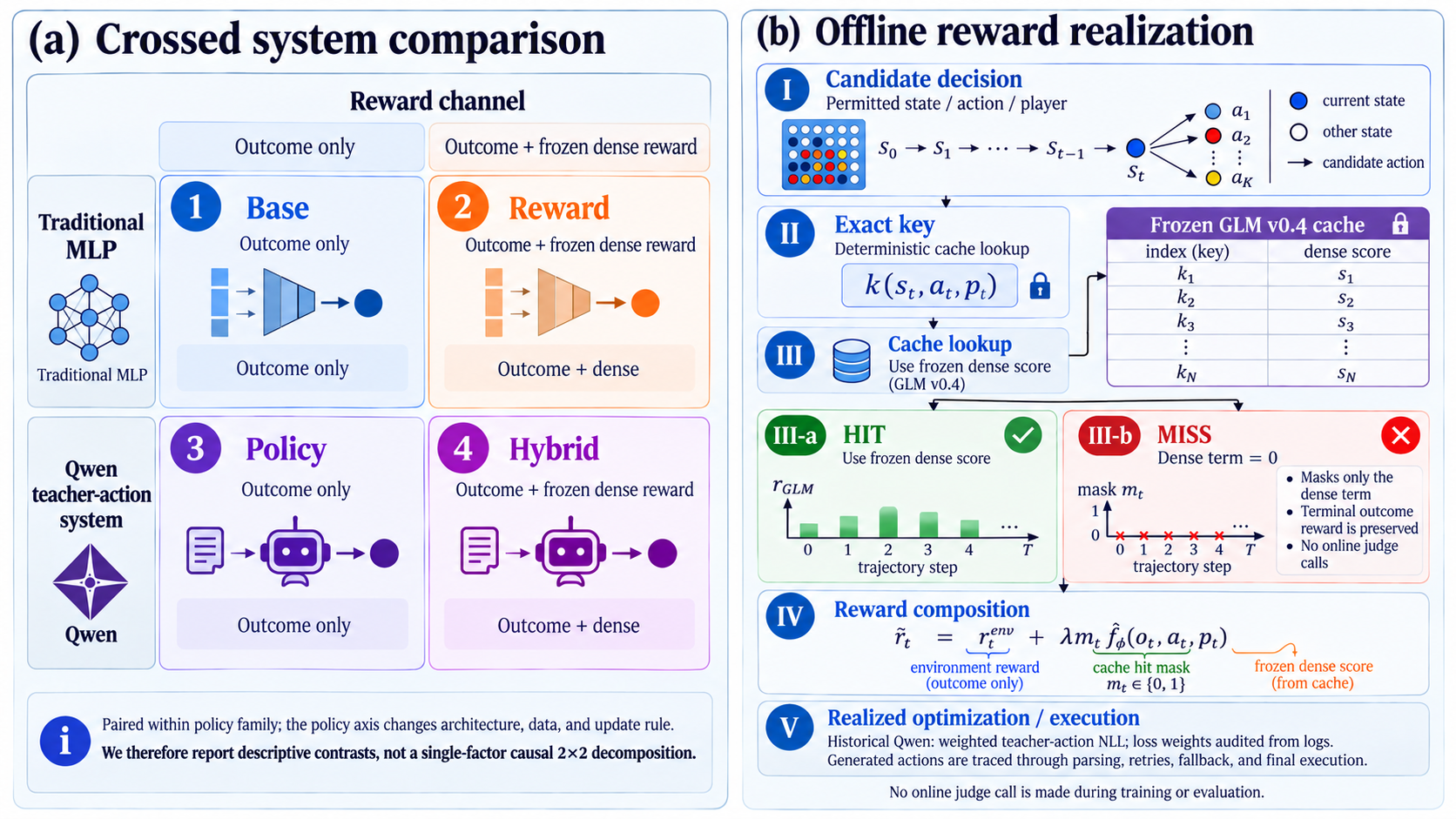}
\caption{Historical Connect Four design. Base--Reward and Policy--Hybrid are within-family contrasts; because the policy family also changes architecture, data, and update rule, Eq.~\ref{eq:interaction} is descriptive rather than a single-factor causal decomposition. Dense reward is an exact frozen-cache lookup; optimization and execution are audited downstream.}
\label{fig:system-overview}
\end{figure}

Let $o_t$ be the permitted observation, $a_t$ the selected legal action, and $p_t$ the acting player. Dense-reward arms use
\begin{equation}
\widetilde r_t=r_t^{\rm env}+I_r\lambda m_t\widehat f_\phi(o_t,a_t,p_t),
\label{eq:reward}
\end{equation}
where $I_r$ enables the dense intervention and $m_t$ records an exact cache hit. A miss masks only the dense term; terminal reward is preserved. Historical arms are Base (traditional policy, outcome-only), Reward (traditional policy, dense reward), Policy (Qwen system, outcome-only), and Hybrid (Qwen system, dense reward). For empirical arm win rates $\mu_B,\mu_R,\mu_P,\mu_H$, we report within-family contrasts and their difference:
\begin{equation}
\Delta_T=\mu_R-\mu_B,\qquad \Delta_Q=\mu_H-\mu_P,\qquad I=\Delta_Q-\Delta_T.
\label{eq:interaction}
\end{equation}
The policy axis changes architecture, training data, and update rule. These quantities therefore describe a crossed system comparison, rather than a single-factor causal decomposition. Figure~\ref{fig:system-overview} shows the historical deployment; Appendix~\ref{app:protocol} specifies the learners and budgets.

A \emph{realization contract} records the intended learner, reward artifact, data source, update rule, action wrapper, checkpoint, evaluation schedule, and information boundary. Starting from an outcome contrast, Table~\ref{tab:main-protocol} asks one falsifiable question at each interface: does the reference mean what we think, does the signal reach updates, is the declared objective numerically realized, does the proposed action survive execution, and what evidence is statistically or historically independent? A failed check triggers a targeted follow-up rather than a post-hoc narrative. Multiple discrepancies may coexist; the protocol does not assume a unique causal explanation.

\begin{table}[!htbp]
\caption{Operational LocusRL checks. Metrics retain their own denominators; a failed check determines the next diagnostic or corrective action rather than being collapsed into one score.}
\label{tab:main-protocol}
\centering\footnotesize
\begin{tabularx}{\linewidth}{>{\raggedright\arraybackslash}p{0.17\linewidth}YY}
\toprule
Check & Quantified object / decision basis & Follow-up action\\
\midrule
Reward semantics & Agreement count / total under a fixed reference; report category and denominator & Recheck the reference, then audit frozen outputs\\
Signal delivery & Cache hits / queries; optimization samples receiving nonzero dense reward & Separate queryability, realized delivery, and nonzero signal\\
Optimization objective & Numerical match between logged loss and declared formula; sampled--update log-probability discrepancy & Verify the implemented objective; distinguish teacher-action learning from on-policy updates\\
Action execution & Proposed, parsed, fallback, and executed actions; replayed terminal state and payoff & Localize the divergence and verify a concrete replacement\\
Evidence validity & Checkpoint--seed pairing; repeated trajectories; historical exposure overlap & Set the statistical unit, re-evaluate under a common contract, or revise validation status\\
\bottomrule
\end{tabularx}
\end{table}

Human judgment enters when formulating the diagnostic hypothesis, defining the reference, and interpreting the result. Once the reference and realization contract are fixed, counting, numerical identity checks, exposure matching, and rule-based replay are script- or environment-executable. LocusRL is therefore human-guided but repeatable, not a fully autonomous auditor. A check may establish an observation, localize it to a seed/path/state/component, discriminate mechanisms under a controlled follow-up, or verify a concrete correction; these are evidence roles rather than claims that every discrepancy has a single cause.

\subsection{Replication and Executable Certificates}
Training seeds are the replication blocks for learning comparisons. Games, opponent seeds, seats, and repeated trajectories remain nested within a trained checkpoint; re-evaluation does not create another training replicate. The initial Connect Four batch has three paired training seeds and the added batch seven. We report their seed-wise contrasts and equal-weight means separately. Exact sign tests use nonzero paired seed contrasts and report ties; Appendix~\ref{app:statistical-diagnostics} gives assumptions and finite-sample resolution.

\begin{figure}[!t]
\centering
\includegraphics[width=0.82\linewidth]{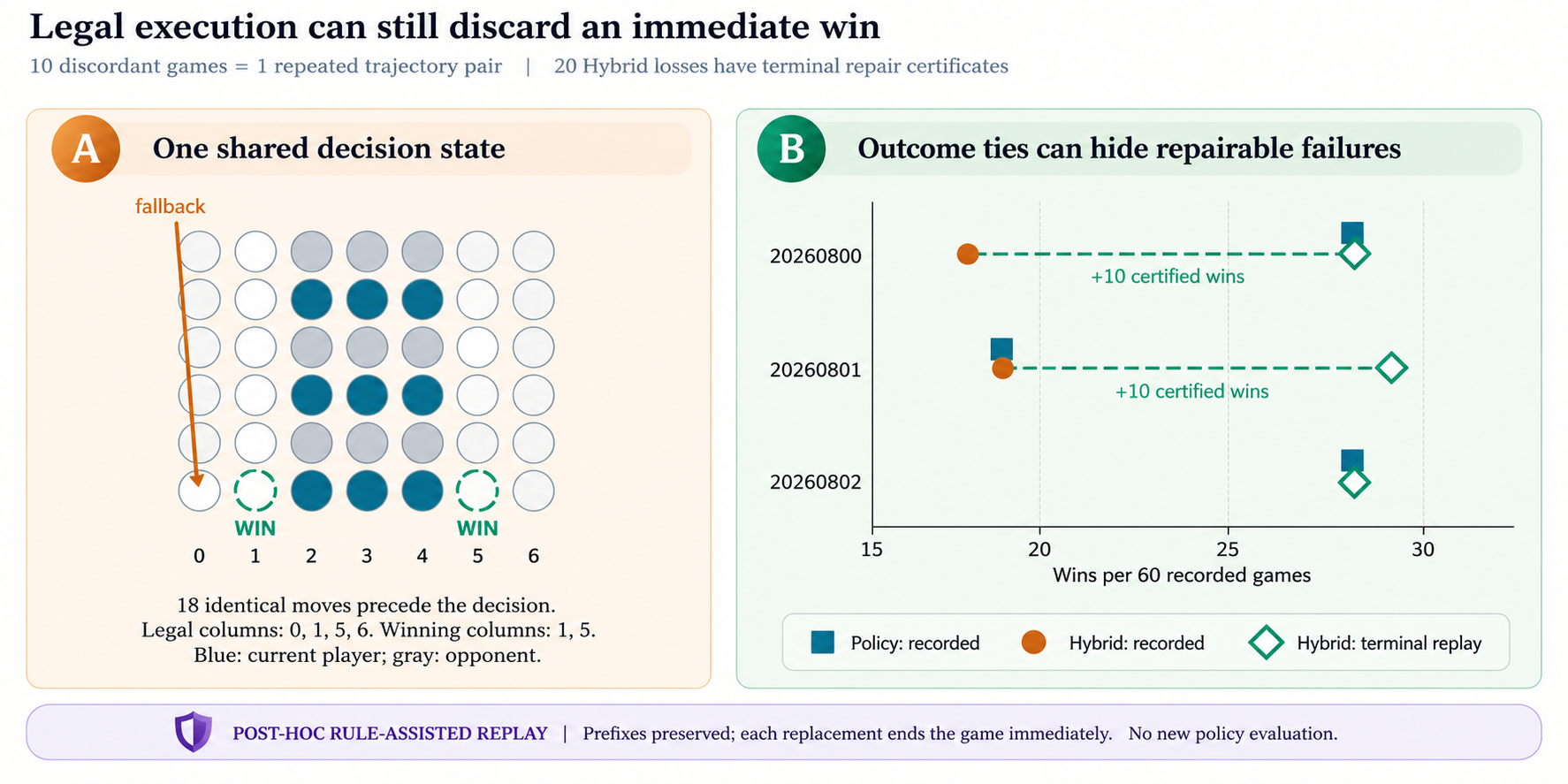}
\caption{Execution diagnosis and terminal replay. The recorded Hybrid fallback misses an immediate win. Open diamonds mark 20 replay-certified repairs that share one first-repair board across two checkpoints; they are not 20 independent tactical situations.}
\label{fig:terminal-replay}
\end{figure}

For terminal action repair, let $W(s)$ contain legal actions that win immediately. At the first recorded fallback with $W(s)\neq\varnothing$, the verifier replays the unchanged prefix, substitutes $a^*\in W(s)$, and checks that the next state is terminal with payoff $+1$ for the acting player. The game ends at that decision, so no subsequent policy response is imputed. Figure~\ref{fig:terminal-replay} illustrates the resulting certificate. Each certificate establishes a sufficient correction at that recorded state; repeated prefixes are counted separately from distinct tactical situations.

\FloatBarrier
\section{Connect Four: Diagnosing and Testing Interventions}
\label{sec:case-initial}

\subsection{Aggregate Contrasts, Seed Variation, and Common Evaluation}
\begin{figure}[!t]
\centering
\includegraphics[width=0.82\linewidth]{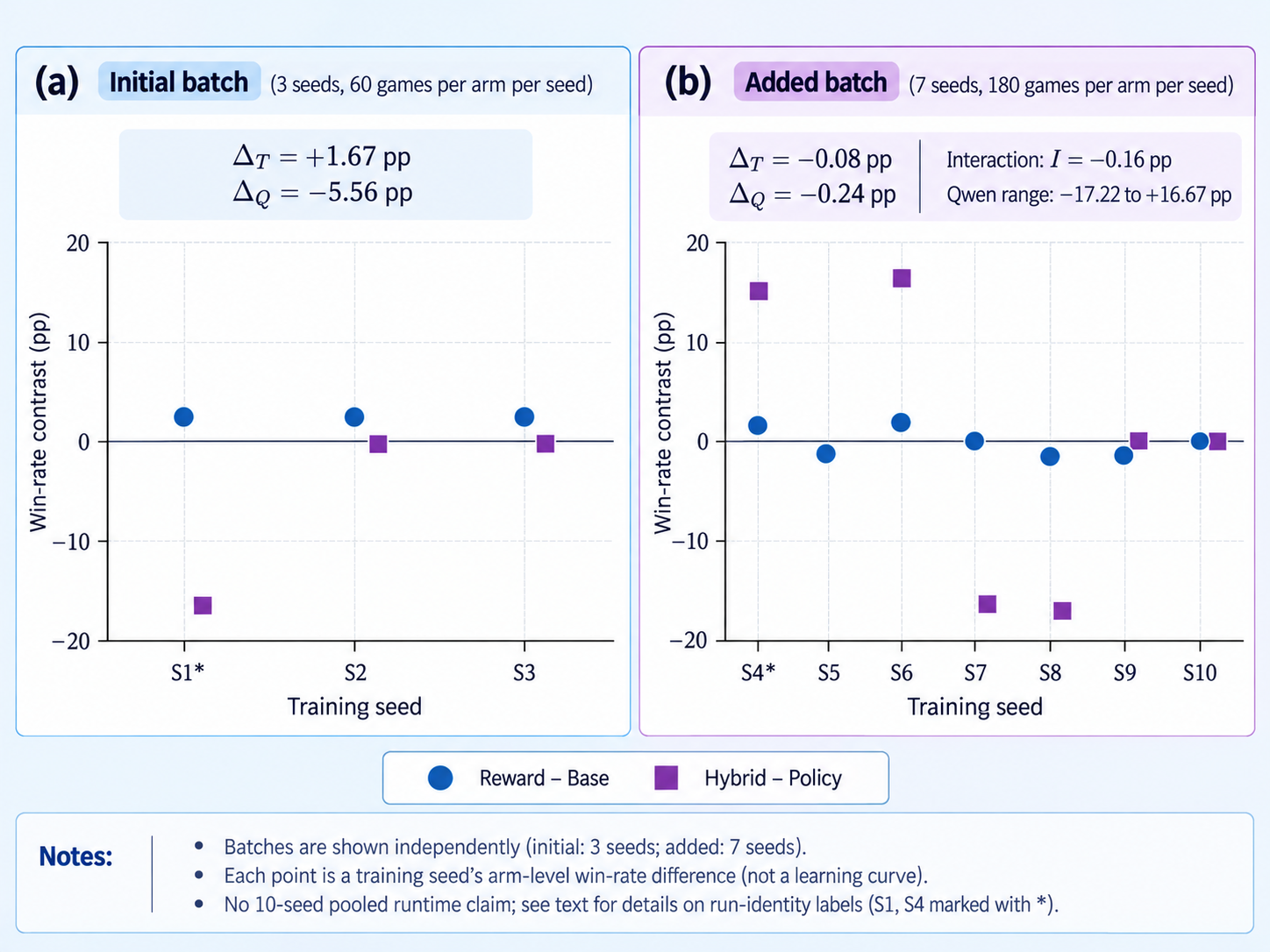}
\caption{Historical seed-wise contrasts, keeping the three-seed and seven-seed batches separate. S1--S10 denote training seeds 20260800--20260809; runtime-label exceptions belong to the historical evaluations. Common-runtime results are reported in Table~\ref{tab:common-runtime}; exact coordinates are in Table~\ref{tab:historical-seed-contrasts}.}
\label{fig:historical-seed-contrasts}
\end{figure}

\textbf{Finding 1---aggregate effects are seed-local and do not identify a stable mechanism.} The initial historical batch contains three training seeds and 720 evaluation games. Wins are Base $93/180$, Reward $96/180$, Policy $75/180$, and Hybrid $65/180$: $\Delta_T=+1.67$ and $\Delta_Q=-5.56$ percentage points (pp), giving interaction $I=-7.22$ pp under Eq.~\ref{eq:interaction} (Table~\ref{tab:paired-inference}). The Qwen deficit is entirely localized to the first seed. Seven additional training seeds add 5,040 games; their Qwen contrasts contain two positive values, two negative values, and three ties, with mean $-0.24$ pp and range $[-17.22,+16.67]$ pp (Figure~\ref{fig:historical-seed-contrasts}; exact values in Table~\ref{tab:historical-seed-contrasts}). The small mean consequently conceals opposing seed-wise outcomes rather than establishing negligible intervention effects.

A common-runtime re-evaluation holds the historical checkpoints from all ten training seeds fixed and applies one implementation and opponent/seat schedule. Table~\ref{tab:common-runtime} reports 800 games, 200 per arm. Reward and Base have equal aggregate wins, while Hybrid records six fewer wins than Policy. Because the schedule also differs from the historical schedules, the change from the old results is not a runtime-only causal estimate. It establishes the behavior of fixed checkpoints under one shared evaluation contract.

\begin{table}[t]
\caption{Common-runtime evaluation of historical checkpoints. Entries are wins; reusing trained checkpoints adds evaluation records, not training replicates.}
\label{tab:common-runtime}
\centering\small
\begin{tabular}{lrrrrr}
\toprule
Training batch & Games/arm & Base & Reward & Policy & Hybrid\\
\midrule
Initial 3 seeds & 60 & 37 & 37 & 34 & 29\\
Added 7 seeds & 140 & 96 & 96 & 97 & 96\\
\midrule
Descriptive total & 200 & 133 & 133 & 131 & 125\\
\bottomrule
\end{tabular}
\end{table}

\subsection{Optimization Identity and an On-Policy Follow-Up}
\label{sec:qwen}
\begin{figure}[!t]
\centering
\includegraphics[width=0.83\linewidth]{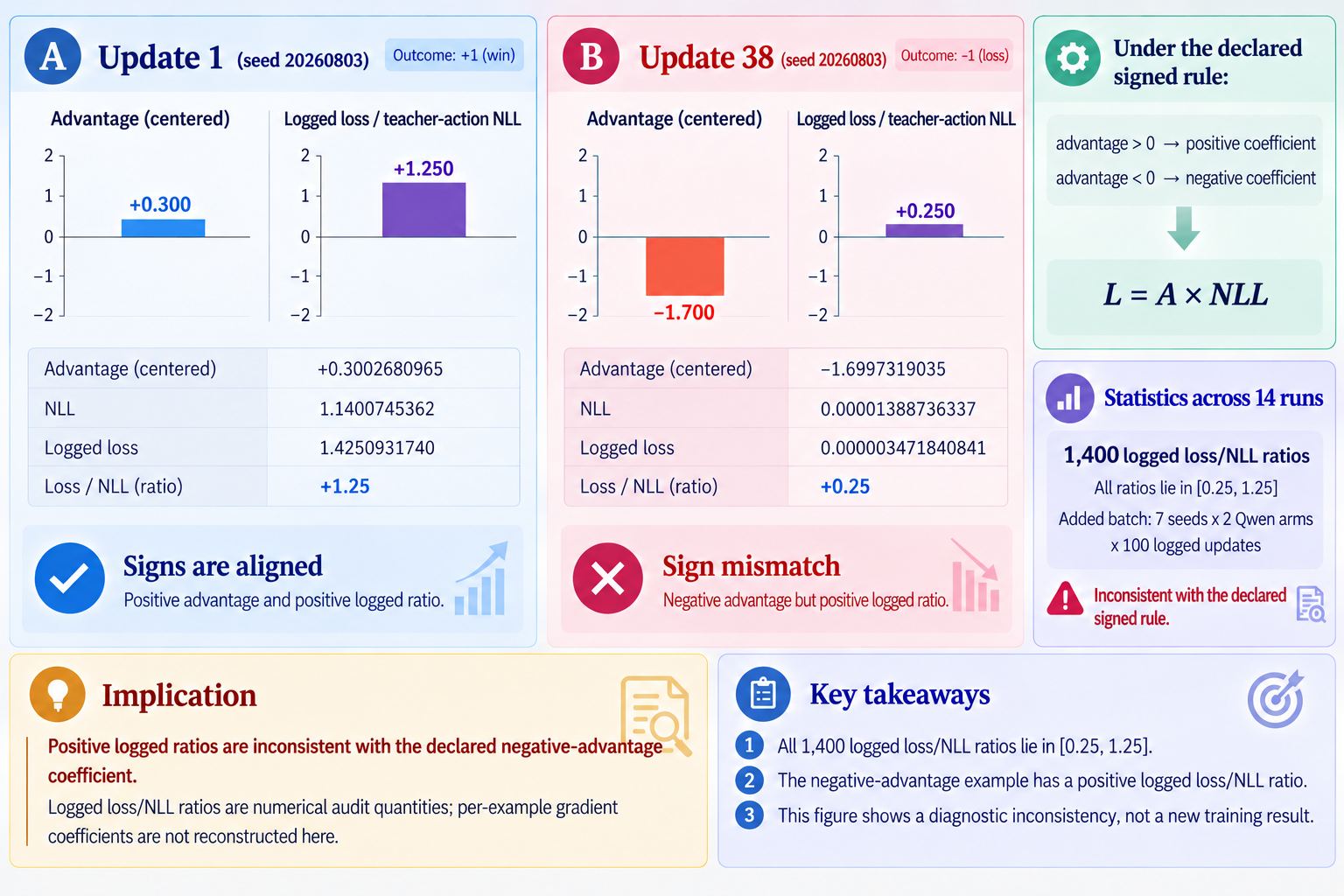}
\caption{Historical numerical objective check. The declared signed rule predicts a negative ratio for negative common advantage, but the archived loss/NLL ratio remains positive. These are logged ratios, not reconstructed per-example gradient coefficients; all 1,400 added-batch ratios lie in $[0.25,1.25]$.}
\label{fig:qwen-update}
\end{figure}

\textbf{Finding 2---the historical Qwen ``policy'' intervention is teacher-action weighted, not verified on-policy.} Teacher-action weighting and self-sampled on-policy learning optimize different objects, so performance comparisons must first identify the realized update. Historical Qwen runs adapt to archived teacher actions with a weighted negative log-likelihood (NLL) form:
\begin{equation}
\mathcal L_{\rm Qwen}=-\frac{1}{|\mathcal D|}\sum_t w_t\log\pi_\theta(a_t^\star\mid o_t,h_t),
\label{eq:qwen}
\end{equation}
where $a_t^\star$ is the teacher action and $h_t$ denotes the supplied history. The historical summary described $w_t$ as a signed centered return. Numerical records contradict that description. For seed 20260803, a batch with advantage $+0.300268$ has loss/NLL $=1.25$, while a constant-outcome batch with advantage $-1.699732$ has the positive ratio $0.25$. All 1,400 logged ratios in the added batch lie in $[0.25,1.25]$ to floating-point tolerance. Thus the recorded objective exhibits positive bounded effective weighting rather than the declared signed coefficient (Figure~\ref{fig:qwen-update}). The historical Policy/Hybrid comparison therefore reflects teacher-action-weighted learning, not a demonstrated self-sampled on-policy update. In mixed batches, the ratio is an NLL-weighted aggregate; it does not identify every per-example coefficient or reconstruct the full historical gradient path.

We instantiate the alternative mechanism in a separate on-policy experiment. Qwen samples from a categorical distribution normalized over legal-action sequence scores, and the update differentiates the same normalized log-probability. Policy and Hybrid each complete three seeds of 200 episodes and 100 updates. The records establish parameter changes, zero sampled/recomputed pre-update log-probability discrepancy, and successful checkpoint reloads. Formal evaluation uses the six original training checkpoints, with 20 games per arm and seed against \texttt{random\_v1} and \texttt{heuristic\_v1}, balanced across seats and paired by environment and action-sampling RNG.

\begin{table}[t]
\caption{Formal evaluation of the new on-policy checkpoints. Each cell has 20 games; there are no draws. The replication blocks are three paired training seeds, with two negative Hybrid--Policy contrasts and one tie.}
\label{tab:onpolicy-formal-results}
\centering\small
\begin{tabular}{lrrr}
\toprule
Training seed & Policy wins & Hybrid wins & Hybrid--Policy\\
\midrule
20260800 & 12/20 & 10/20 & $-10.0$ pp\\
20260801 & 15/20 & 11/20 & $-20.0$ pp\\
20260802 & 12/20 & 12/20 & $0.0$ pp\\
\midrule
Descriptive total & 39/60 & 33/60 & $-10.0$ pp\\
\bottomrule
\end{tabular}
\end{table}

Policy wins $39/60$ games and Hybrid $33/60$ (Table~\ref{tab:onpolicy-formal-results}). This resolves \emph{mechanism identity}: the new checkpoints satisfy an explicit on-policy contract, yet the three inspected Hybrid--Policy contrasts remain negative or tied. It does not establish a universal dense-reward penalty, nor a budget-matched advantage from replacing the historical objective. A plain teacher-action SFT reference adds an unweighted mechanism: 80 examples and 10 optimizer steps per seed yield $66/180$ wins on the historical three-opponent subset and $66/240$ including alpha-beta-d6. Because initial-batch sample and update budgets are not fully recoverable, SFT remains a mechanism reference rather than a strict ablation. Appendix~\ref{app:training-references} gives the contracts and checkpoint identities.

\subsection{Reward Influence Versus Reward Benefit}
\textbf{Finding 3---reward influence is experimentally separable from reward benefit.} A distinct three-seed control changes only the sign of a historical rule-based dense reward while preserving terminal reward and matched initialization. At step 100, sign reversal produces severe degradation in every seed, whereas the original-sign reward does not exceed terminal-only learning (Table~\ref{tab:direction-control}). The result separates sensitivity to a learning signal from its usefulness: the learner responds strongly to reward direction, yet adding the original-sign signal does not improve performance in this control.

\begin{table}[t]
\caption{Step-100 reward-direction control: wins out of 60 per seed. Positive denotes the original sign of the historical rule oracle; the oracle is not assumed tactically perfect under the revised reference.}
\label{tab:direction-control}
\centering\small
\begin{tabular}{lrrr}
\toprule
Training seed & Terminal only & Positive oracle & Sign-reversed oracle\\
\midrule
20260800 & 38 & 30 & 1\\
20260801 & 27 & 27 & 1\\
20260802 & 38 & 27 & 3\\
\midrule
Total & 103/180 & 84/180 & 5/180\\
Win rate & 57.2\% & 46.7\% & 2.8\%\\
\bottomrule
\end{tabular}
\end{table}

The manipulated signal is a rule reward, not the frozen GLM cache. This controlled sign test therefore complements the historical system comparisons without attributing their differences to the same mechanism. Appendix~\ref{app:direction100} reports learning curves, nonzero signal exposure, and schedule details.

\begin{figure}[!t]
  \centering
  \includegraphics[width=0.87\linewidth]{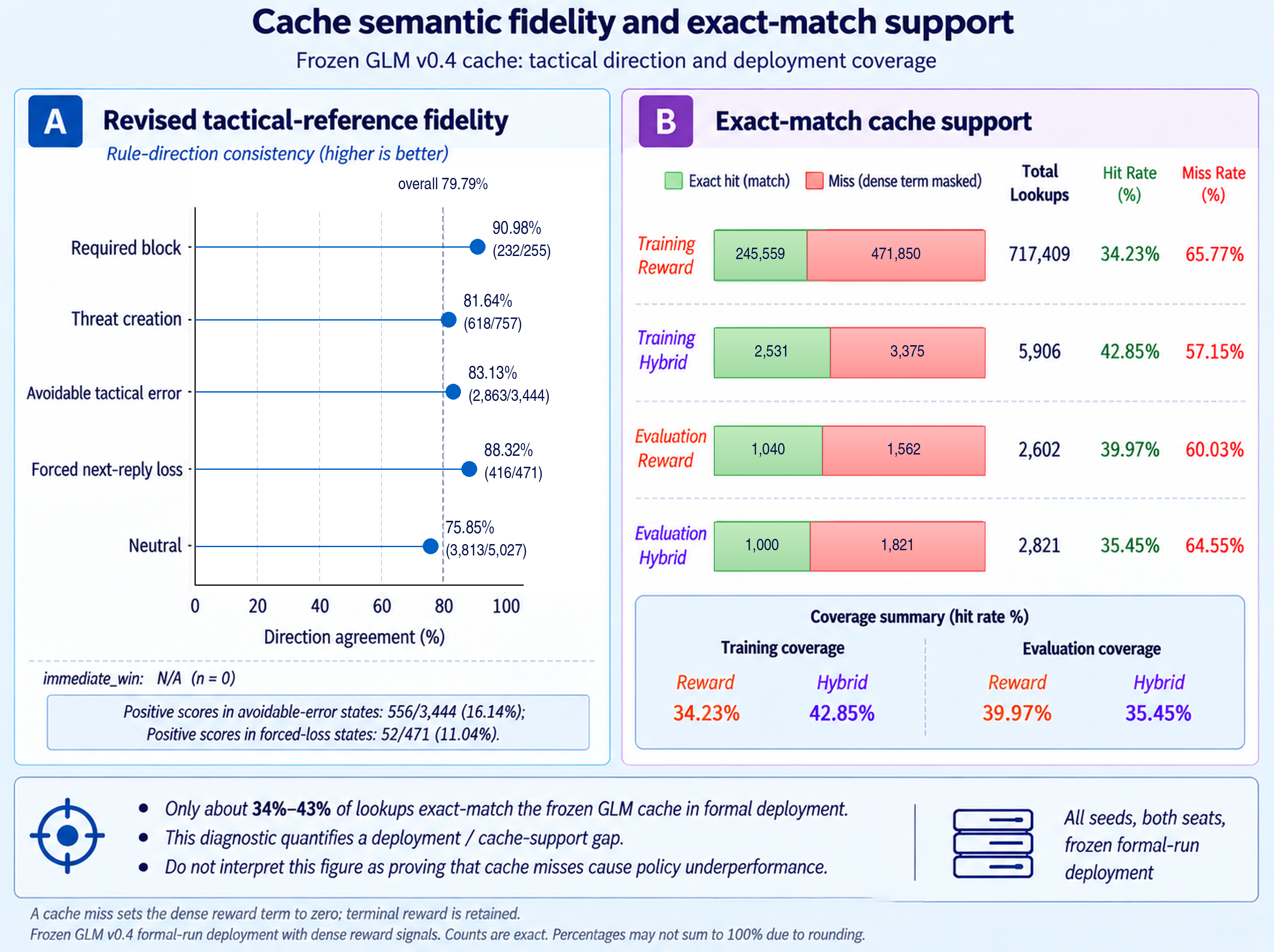}
  \caption{Frozen-cache audit under the revised reference. Left: category-wise semantic fidelity; right: exact-match support under separate training/evaluation lookup denominators. Forced-loss agreement is $88.32\%$ and positive avoidable-error scores total $556/3{,}444$; Table~\ref{tab:glm-cache-audit} gives all counts.}
  \label{fig:cache-fidelity-coverage}
\end{figure}

\subsection{Reward Semantics and Realized Support}
\textbf{Finding 4---structural validity, semantic fidelity, delivery, and utility are distinct.} Reward quality is not a single scalar property: semantic correctness, realized delivery, influence on learning, and performance benefit require separate checks. For semantics, all 9,954 frozen cache entries pass structural integrity checks, but successor enumeration exposes errors in the original tactical reference. The revised categories contain 255 required blocks, 757 threat creations, 3,444 avoidable tactical errors, 471 forced next-reply losses, and 5,027 neutral entries; direction agreement is $7{,}942/9{,}954=79.79\%$. Positive frozen scores occur for 556 avoidable errors and 52 forced losses, motivating reference correction while leaving the stored GLM scores unchanged (Appendix~\ref{app:reference-revalidation}).

Delivery then asks a different question. In the initial historical batch, exact training-cache hit rates are $34.23\%$ for Reward and $42.85\%$ for Hybrid under their respective lookup denominators; a hit may still return zero or a directionally inconsistent score. Finally, the independent reward-direction control separately tests influence versus benefit: terminal-only, positive-oracle, and sign-reversed win rates are $57.2\%$, $46.7\%$, and $2.8\%$. Because these checks use different objects and experimental sources, their denominators remain separate and are not multiplied into an ``effective reward rate.'' Figure~\ref{fig:cache-fidelity-coverage} and Appendix~\ref{app:cache-audit} report the support and fidelity counts.

\subsection{From Legal Execution to a Verified Winning Correction}
\textbf{Finding 5---legal execution can still be tactically wrong, and the local repair is replay-certifiable.} Action tracing identifies a distinct failure. In the discordant Qwen pair, the first 18 moves are identical. Legal columns are $\{0,1,5,6\}$, with 1 and 5 immediate wins. Policy plays 1; Hybrid proposes a full column, exhausts three attempts, and falls back to 0 before losing. Thus final legality hides a missed terminal win.

The same board appears under a second Hybrid checkpoint. Applying the same fallback-only immediate-win rule symmetrically certifies 20 recorded Hybrid loss-to-win changes and no Policy changes (Figure~\ref{fig:terminal-replay}). These records share one first-repair board across two checkpoints. Replaying the unchanged prefixes and checking immediate terminal payoffs verifies a sufficient local correction without imputing later behavior. The result closes the execution diagnosis at that state; it is separate from prospective evaluation of a deployed wrapper.

\section{Imperfect-Information Extensions}
\label{sec:leduc-outcomes}
Leduc and Goofspiel stress the same realization interfaces under hidden information, equilibrium-dependent reward references, and validation history. The purpose is not to claim game-independent effect sizes, but to test whether the diagnostic questions still produce concrete decisions. The reward gates use explicit accuracy, request-failure, and leakage criteria; all three confirmation results stop formal release despite meeting the failure-rate and leakage criteria (Table~\ref{tab:main-gates}).

\begin{table}[!htbp]
\caption{Reward-release decisions under the recorded gate criteria. Accuracy retains every registered pair; failures count action requests. Historical freeze and exposure evidence is detailed separately in Appendices~\ref{app:leduc-gate} and~\ref{app:goofspiel-gate}.}
\label{tab:main-gates}
\centering\small
\begin{tabular}{lrrrrl}
\toprule
Reward & Correct & Accuracy & Failures & Leakage & Formal cache\\
\midrule
Leduc v0.1 & 9/24 & 37.5\% & 1/48 & 0 & Stopped\\
Goofspiel v0.1 & 7/29 & 24.1\% & 0/58 & 0 & Stopped\\
Goofspiel v0.2 & 18/29 & 62.1\% & 2/58 & 0 & Stopped\\
\bottomrule
\end{tabular}
\end{table}

\paragraph{Exposure and reference-target checks.}
Leduc v0.2 records $20/24$ on currently disjoint splits, but all 24 confirmation states had appeared in the opened v0.1 development set; this is re-validation, not independent confirmation. In Goofspiel, 47 of 90 exposed v0.3 development pairs preserve direction across the audited equilibrium-opponent mixtures, 42 can become ties, and one can reverse. Thus equilibrium selection is part of the reference target, and calibration must specify a selected opponent mixture or a direction robust across admissible mixtures. The later v0.4 delivery reports $64/90$ in development and stops with confirmation unopened.

\paragraph{Training and behavior under a fresh evaluation schedule.}
Outcome-only Leduc training evaluates three paired Base/Policy seeds over 180 games per arm under a fresh RNG schedule. Base and Policy average $0.7778$ and $0.6889$ chips/game, with seed differences $+0.0667$, $-0.4833$, and $+0.1500$. On visited evaluation states, two Policy checkpoints always raise when legal while the third always checks/calls, despite successful update, probability-consistency, and reload checks. The traces reveal strategies hidden by aggregate returns without assigning the return differences to one pattern; Appendix~\ref{app:leduc-policy} gives the full contract.

\section{Discussion and Limitations}
\label{sec:synthesis}
\paragraph{What the protocol changes.} The central shift is to evaluate the \emph{realized intervention path}, not merely the declared component or its aggregate score. Three lessons follow. \textbf{First, component validity is not system utility:} structurally valid rewards can fail semantic checks, cache access need not yield useful signal, and learning influence need not improve performance. \textbf{Second, the realized mechanism is part of the treatment:} teacher-action likelihood differs from on-policy updates, and legal fallback can replace a winning action. \textbf{Third, evidence units matter:} means can hide seed reversals, repeated games can share one failure state, and currently disjoint splits can reuse exposed examples. Practically, a negative arm contrast should not immediately trigger reward redesign: the protocol first asks whether the reward was semantically valid, delivered, optimized as declared, and executed as intended.

\paragraph{Transfer conditions.} The protocol maps reference counterexamples to successor checks, loss mismatches to objective verification, execution divergences to replay, and provenance uncertainty to fixed-checkpoint re-evaluation. Transfer requires matching these interfaces: algorithm-specific loss/update records for optimization, the reward path for delivery, and proposed plus executed actions for action checks. Offline RL additionally needs data-source/coverage provenance; multi-agent RL needs opponent versions and information boundaries. Exact terminal replay requires recoverable state and executable rules; otherwise repairs need repeated or statistical validation. Evidence here remains limited to Connect Four, Leduc, and Goofspiel (Table~\ref{tab:diagnostic-yield}).

\paragraph{Scope of inference.}
The historical policy families differ in architecture, data, and update rule, and training replication is modest. Replay certificates cover one first-repair state across two checkpoints; Leduc adds three seeds against one opponent family. Initial Qwen records do not recover strict SFT budget matching; the added batch records 1,400 updates, but the full return-to-weight/backpropagation mapping remains unresolved. These limits constrain causal and generalization claims, not the directly checkable numerical contradictions, fixed-checkpoint outcomes, or terminal certificates.

Prospective diagnostic recall, false-positive rate, and human-time savings on new systems remain unmeasured. The present evidence establishes a human-guided, repeatable procedure; measuring those meta-diagnostic quantities is future work.

\section{Conclusion}
LocusRL changes the object of evaluation from a declared LLM component to its realized intervention path: what reward is delivered, what objective is optimized, what action is executed, and what evidence supports the comparison. Across competitive-game experiments, this shift exposes seed-local effects, distinguishes teacher-action weighting from verified on-policy learning, separates reward influence from reward benefit, and certifies a local execution repair. The result is not a universal claim about dense rewards; it is an operational way to turn ambiguous score differences into testable mechanism claims and concrete decisions about training, evaluation, and repair.

\section*{Reproducibility Statement}
Code and selected evidence: \url{https://github.com/luancy0423/LocusRL-Pub}.
Appendix~\ref{app:protocol} specifies training/evaluation contracts and distinguishes configured, completed, and observed budgets. Appendices~\ref{app:runtime-identity}, \ref{app:statistical-diagnostics}, and~\ref{app:terminal-replay} document runtime identities, replication units, and replay certificates. Appendix~\ref{app:training-references} records SFT and the six-checkpoint, 120-game formal on-policy evaluation, including checkpoint correspondence, paired RNG schedules, and artifact verification. Appendices~\ref{app:reference-revalidation}, \ref{app:direction100}, and~\ref{app:leduc-policy} supply reference reanalysis, reward-direction controls, and Leduc reproduction details. The included \texttt{verify\_replication\_units.py} reproduces frozen historical seed-summary arithmetic. Historical recovery limits and the distinction between lightweight evidence delivery and model-weight availability are specified in Appendix~\ref{app:verification-targets}. The accompanying selected code/evidence release provides CPU-only checks via \texttt{verify\_results.py --full}; its README lists included records and omitted inputs. Model weights, the complete historical replay archive, and some historical training inputs are not part of that release.

\section*{AI Use Disclosure}
Generative AI tools assisted with research methodology feedback, code discussion and implementation support, data cleaning/reformatting, qualitative analysis, interpretation, literature organization, figure editing, and manuscript drafting. Language models also serve as experimental policy/reward components. For the 60-case reference review, manual labels preceded AI-assisted second-stage checking; a later 30-case supplement used AI first-pass analysis followed by human case-by-case confirmation. These workflows are reported separately. The authors reviewed the AI-assisted outputs used in the paper and take responsibility for its text, code, figures, and empirical claims.

\bibliographystyle{iclr2027_conference}
\bibliography{iclr2027_conference}
\clearpage
\appendix

\FloatBarrier
\section{Prior Work by Intervention Axis}
\label{app:prior-work-map}

Table~\ref{tab:prior-axis} maps component-learning methods by intervention.
Table~\ref{tab:related-audits} compares the closest audit methods by their
object, evidence, and validated intervention. These are complementary
scientific targets: LocusRL investigates how a supplied reward is realized
inside a composed system, and verifies a local execution repair.

\begin{table}[ht]
\caption{Closest audit methods and their relation to the present contribution.
A local terminal replay and a policy-level repair are different validation units.}
\label{tab:related-audits}
\centering\footnotesize
\begin{tabularx}{\linewidth}{p{0.18\linewidth}YYY}
\toprule
Method & Audit object & Evidence or estimator & Intervention / validation\\
\midrule
Alignment Auditor \citep{bou2026alignment} & Implicit LLM objectives & Bayesian IRL and posterior uncertainty & Inferred-reward refinement and downstream RLHF validation\\
Hidden-objective audits \citep{marks2025auditing} & Undesired internal objective & Behavioral, interpretability, and training-data investigations & Blind auditing exercise on a model with a hidden objective\\
IR$^3$ \citep{beigi2026ir3} & Reward-hacking features of implicit objectives & Contrastive IRL and feature decomposition & Targeted mitigation with behavior/capability evaluation\\
Spurious Rewards \citep{shao2025spurious} & RLVR reward informativeness & Alternative reward signals across model families & Measured training responses to spurious rewards\\
RATE \citep{reber2025rate} & Reward-score attribute sensitivity & Rewrites with counterfactual bias correction & Attribute-effect estimation under stated assumptions\\
LocusRL & Realized LLM interventions in composed learners & System contrasts, objective and support logs, action traces, and validation histories & Controlled follow-ups, terminal-replay correction, and reward-release decisions\\
\bottomrule
\end{tabularx}
\end{table}

\begin{table}[ht]
\caption{Representative LLM-for-RL work organized by the intervention it
identifies. The final column states what is still needed to test the
composed-system diagnostic question studied by LocusRL.}
\label{tab:prior-axis}
\centering
\footnotesize
\begin{tabularx}{\textwidth}{
  >{\raggedright\arraybackslash}p{0.19\textwidth}
  >{\raggedright\arraybackslash}p{0.17\textwidth}YY}
\toprule
Line of work & LLM locus & Primary question identified &
Relation to system comparison\\
\midrule
ELLM; Reward Design; Motif
\citep{du2023ellm,kwon2023rewarddesign,klissarov2024motif}
& Goal proposal, proxy reward, or preference-derived intrinsic reward
& Can language-model knowledge improve exploration or replace costly reward
engineering?
& Whether the same reward artifact retains its effect after replacing the
downstream policy family.\\
\addlinespace
Text2Reward; Eureka
\citep{xie2024text2reward,ma2024eureka}
& Executable dense-reward synthesis and refinement
& Can an LLM generate reward code whose induced policy achieves high return?
& Reward presence is not crossed with conventional versus LLM policy systems
using one frozen artifact.\\
\addlinespace
GLAM; RAGEN; SPIRAL
\citep{carta2023glam,ragen2025,spiral}
& LLM policy or self-play policy
& Can language policies be grounded and stably optimized through interaction?
& The reward channel is not independently toggled under matched controls, and
generation failures become part of the policy treatment.\\
\addlinespace
RLAnything \citep{rlanything}
& Jointly adapting policy, reward model, and environment
& Do mutually adapting components improve an end-to-end LLM-agent system?
& Co-adaptation is the method; it does not estimate the transfer of one frozen
reward across policy families.\\
\addlinespace
Policy-invariant LLM shaping \citep{hounwanou2026policyinvariant}
& LLM planner and potential-based shaping
& Can LLM feedback be incorporated without changing the optimal-policy set?
& The theoretical guarantee does not estimate empirical interaction with an
LLM policy; reported training evidence is a small pipeline validation.\\
\addlinespace
\textbf{LocusRL}
& \textbf{Traditional/LLM policy $\times$ absent/present frozen LLM reward}
& \textbf{Does the identical reward compose across policy systems, and is the
realized treatment semantically and operationally valid?}
& \textbf{Connect Four supplies system comparisons and controls; Leduc adds
training/evaluation evidence, and Goofspiel adds reward-reference and
validation diagnostics.}\\
\bottomrule
\end{tabularx}
\end{table}

\FloatBarrier
\section{Reward Calibration Details}
\label{app:calibration}

\subsection{Reachability, Paired Scoring, and Selection Rules}

The Connect Four calibration set consists of 80 preferred/dispreferred action
pairs at reachable positions, balanced across both acting-player seats and
four tactical categories. Reachability and tactical labels serve different
checks: replay establishes that a position and its acting player are valid;
the rule oracle orders candidate actions at that position. Category-mean
scores do not replace this within-position ordering test.

Let $\mathcal V$ contain the registered pairs with two valid scores. The
strict scorable-pair accuracy is
\begin{equation}
 \operatorname{Acc}_{\rm scorable}=\frac{1}{|\mathcal V|}
 \sum_{i\in\mathcal V}\mathbf{1}\!\bigl[
 \widehat f_\phi(s_i,a_i^+,p_i)>\widehat f_\phi(s_i,a_i^-,p_i)\bigr],
 \label{eq:pairwise}
\end{equation}
where ties count as incorrect. Conservative accuracy uses the same correct
count divided by all 80 registered pairs, retaining exhausted requests in the
denominator. These definitions distinguish directional accuracy from API
availability. The candidate-judge reference requires at least 85\% scorable
accuracy and 80\% conservative accuracy; the later imperfect-information
gates use their own 75\% accuracy, 5\% failure, and zero-leakage criteria.

The historical reasoning-enabled DeepSeek confirmation, first-exposure GLM
screen, and subsequent GLM must-block revision are separate stages. The
revision is development on opened states, not another blind confirmation.
The later low-token DeepSeek cache also differs from its reasoning-enabled
confirmation configuration. These distinctions preserve which model/prompt
configuration each score actually evaluates.

\subsection{Tactical Labels and Recorded Configuration Results}

Reachability is established by replay from the empty board with gravity,
alternating turns, legal-column checks, acting-player consistency, and no
earlier terminal state. The historical oracle assigns $+1.0$ to an immediate
win, $+0.8$ to a fork, $+0.6$ to a single threat, $+0.3$ to a required block,
and $0$ to a neutral move; missed wins and blocks receive negative scores.
These are reference labels, not the deployed LLM scores. The calibration
results below retain their original reference and exposure status.
Appendix~\ref{app:reference-revalidation} subsequently corrects the defensive
predicate and separates next-reply forced losses in the 9,954-entry GLM cache;
this reanalysis is not a new confirmation of the historical 80-pair screen.

\begin{table}[ht]
\caption{Historical reasoning-enabled DeepSeek confirmation. Correct counts
are conditional on two valid scores; failed pairs remain in the conservative
denominator.}
\label{tab:confirmation}
\centering
\small
\begin{tabular}{lrrrr}
\toprule
Category & Seat 0 & Seat 1 & Overall & Failed\\
\midrule
Immediate win & $8/10$ & $9/10$ & $17/20$ & $0$\\
Required block & $7/8$ & $7/8$ & $14/16$ & $4$\\
Threat creation & $7/8$ & $7/9$ & $14/17$ & $3$\\
Good move & $10/10$ & $10/10$ & $20/20$ & $0$\\
\midrule
Overall & $32/36$ & $33/37$ & $65/73$ & $7$\\
\bottomrule
\end{tabular}
\end{table}

The scorable-pair accuracy is $89.04\%$ and conservative accuracy is
$81.25\%$. Mean request latency is $192.7$ seconds and the 95th percentile is
$709.7$ seconds; the recorded cost proxy is approximately \$1.46. This result
uses thinking and 4,096--32,768-token escalation and therefore does not
validate low-token V2. On the later frozen screen, first-exposure GLM reaches
$71/76=93.42\%$ scorable and $71/80=88.75\%$ conservative accuracy. The
post-hoc must-block revision reaches $70/76=92.11\%$ and $70/80=87.50\%$,
with $20/20$ required-block preferences. Across the first GLM screen, four
pairs fail, no parser failures occur, and reported usage is 164,486 tokens.

\begin{table}[ht]
\caption{Real-API reward-judge configuration screen on the frozen 80-pair
dual-seat set. ``Scorable'' excludes exhausted API pairs but treats ties as
incorrect. Conservative accuracy counts all failures in the denominator.}
\label{tab:judge-screen}
\centering
\scriptsize
\setlength{\tabcolsep}{4pt}
\begin{tabular}{lrrrrrr}
\toprule
Configuration & Correct/scorable & Scorable acc. & Conservative & Ties & Reversed & Failed\\
\midrule
DeepSeek v4 Flash, low-token & $26/80$ & $32.50\%$ & $32.50\%$ & 25 & 29 & 0\\
Mimo 2.5 Pro, high, 8192 & $35/80$ & $43.75\%$ & $43.75\%$ & 27 & 18 & 0\\
GLM-5.3-Flash, high & $71/76$ & $93.42\%$ & $88.75\%$ & 2 & 3 & 4\\
\bottomrule
\end{tabular}
\end{table}

\FloatBarrier
\section{Experimental Protocol and Reporting}
\label{app:protocol}

\begin{table}[t]
\caption{Implemented Connect Four learning systems and budget evidence. Historical Qwen rows separate what is verified from what remains recoverable.}
\label{tab:learner-contract}
\centering\footnotesize
\begin{tabularx}{\linewidth}{p{0.22\linewidth}YYY}
\toprule
 & Base / Reward & Initial Policy / Hybrid & Added Policy / Hybrid\\
\midrule
Backbone & Masked-action MLP & Qwen2.5-3B-Instruct & Qwen2.5-3B-Instruct\\
Training actions & Learner rollouts & Heuristic teacher actions & Heuristic teacher actions\\
Configured / observed budget & 100 updates $\times$ 200 episodes & \texttt{TRAIN\_MAX\_STEPS}\allowbreak\texttt{=10} configured; completed updates and consumed-example order unavailable & 100 logged updates/run; 1,400 updates across 14 runs\\
Objective & Centered-return policy gradient & Weighted teacher-action NLL; exact historical weight mapping not recovered & Positive bounded loss/NLL ratios verified numerically\\
Dense intervention & Reward adds frozen score & Hybrid adds frozen score to archived return signal & Same frozen reward mechanism\\
\bottomrule
\end{tabularx}
\end{table}

\begin{table}[t]
\caption{Diagnostic workflow. Each audit produces an inspectable object and
specifies the next action when its check does not close the question.}
\label{tab:diagnostic-workflow}
\centering\footnotesize
\begin{tabularx}{\linewidth}{p{0.16\linewidth}YY}
\toprule
Audit layer & Evidence object & Decision or next intervention\\
\midrule
Execution identity & Train/eval source matrix; actual loss versus declared formula & Reconcile source and logging; hold weights fixed for common-runtime evaluation\\
Reward semantics & Reachable action pairs; executable reference counterexamples; full-cache labels & Validate the reference, then distinguish intact bytes from valid scores\\
Realized support & Lookup, nonzero signal, and direction-consistent exposure counts & Determine whether the nominal reward reaches the learner's states\\
Action execution & Proposed/parsed/executed actions; first divergence; terminal certificate & Identify a concrete wrapper repair; freeze it for prospective testing\\
Multiplicity & Paired trajectories and training-seed contrasts & Separate repeated paths from independent retraining evidence\\
Exposure history & Historical opened-state union; frozen splits and stop record & Keep confirmation independent; stop before formal cache or training\\
\bottomrule
\end{tabularx}
\end{table}

\begin{table}[ht]
\caption{Environment and evidence scope. Adapter availability does not imply
completed policy-performance experiments.}
\label{tab:environments}
\centering
\footnotesize
\begin{tabularx}{\linewidth}{lYccY}
\toprule
Environment & Protocol & Max steps & Return & Status\\
\midrule
Connect Four & Perfect information; columns $0,\ldots,6$ & 42 & $-1,0,+1$ & Four-arm results complete\\
Leduc-style poker & Hidden cards; simplified betting; no re-raise & 6 & $[-7,+7]$ & Reward gate stopped; terminal-reward Base/Policy training and 360-game evaluation complete\\
Goofspiel & Public remaining bids; pending bid and future prize order hidden & 10 & $-1,0,+1$ & Validity gate failed; stopped before training\\
\bottomrule
\end{tabularx}
\end{table}

The initial v0.4 protocol uses training seeds 20260800--20260802 and evaluation seeds
20260810--20260812. Each independently trained arm plays ten games in each of
$3$ opponents $\times\,2$ seats, yielding 60 games per seed and 180 per arm.
Base and Reward have verified 100-step, 200-episode-per-step training records.
For the initial Policy/Hybrid runs, the recovered configuration requests
\texttt{TRAIN\_MAX\_STEPS=10}; that field is not treated as proof that ten
updates completed. Frozen teacher-action pools contain 1,125, 1,120, and 1,124
examples by seed, but the original training-instance archive needed to recover
actual completed updates, sample-consumption order, and examples consumed per
checkpoint remains outstanding. The added seven-seed Qwen batch is different:
it contains 100 logged updates per run, yielding 1,400 logged updates across
14 Policy/Hybrid runs.

Reporting keeps strategic outcomes, first-pass invalidity, invalid attempts,
parser failures, retries, fallback, final legality, cache hits/misses, reward
masking, API calls, tokens, latency, update counts, checkpoint identity,
commits, and run IDs as distinct quantities. The 720-game delivery contains
all per-game episode records and its outer archive hash begins
\texttt{B92F9D2E}; 59 listed internal hashes reproduce after normalizing the
archived absolute paths to bundle-relative paths.

\subsection{Environment Adapter and Information Contract}

Every environment exposes acting-player observation, textual state, a fixed
action vocabulary, a legal-action tuple, a binary legality mask, a terminal
flag, and zero-sum returns. The \texttt{GameState} mask has the vocabulary's
length, contains only binary values, and recovers exactly the declared legal
actions. Every nonterminal state admits a legal action. The shared
\texttt{GameAdapter} gives the four arms the same policy-facing interface;
source-identity checks separately establish whether runs executed the same
implementation.

In the sequential Goofspiel adapter, both players' visible hands are those
from the beginning of the round. The second player's observation and public
history exclude the first player's pending bid; displaying a post-bid hand
would reveal the removed card. Future prize order is also hidden. Observation-
and history-level regression checks enforce these boundaries, which must
also hold in reward prompts and cache-key construction. Leduc prompts expose
only the acting player's permitted card and betting information. These are
requirements of the repository protocols in Table~\ref{tab:environments}.

\subsection{Learning Rules and Teacher Data}

The traditional Base and Reward arms share the masked-action backbone,
optimizer configuration, and initialization. Their implemented centered-return
policy-gradient objective is
\begin{equation}
 \mathcal L_{\rm traditional}(\theta)
 =-\frac{1}{|\mathcal D|}\sum_{(o_t,a_t)\in\mathcal D}
 \widehat A_t\log\pi_\theta(a_t\mid o_t),
 \label{eq:traditional}
\end{equation}
with actions collected from the learner. The intended Base/Reward treatment
difference is whether the frozen dense term contributes to the return used
in $\widehat A_t$.

Qwen starts from \texttt{Qwen2.5-3B-Instruct} and uses parameter-efficient
teacher-action adaptation. The supplied implementation describes
\texttt{HeuristicPolicy} demonstrations against a legal-random opponent;
the frozen source manifest does not establish an alpha-beta-d6 teacher. The frozen initial Policy/Hybrid teacher-action pools contain 1,125, 1,120,
and 1,124 examples for seeds 20260800, 20260801, and 20260802. These are
available source examples, not proof that every example was consumed by the
historical optimizer. The original training-instance logs needed to recover
actual sample order and completed updates remain unavailable. The examples are
teacher-action records rather than Qwen-generated on-policy rollouts. Equation~\ref{eq:qwen}
separates weighted NLL from the historical centered-return description of its
weights, which the numerical audit has corrected. Appendix~\ref{app:added-seeds} records why the executed mapping from
returns to weights remains unresolved.

Among the initial paired examples, 1,375 have nonzero dense return. Adding
centered dense return to the archived outcome coefficient changes magnitude
without changing coefficient signs, with within-seed cosine similarities
0.942--0.948. Appendix~\ref{app:mechanism-controls} gives the fixed-data
negation calculation. This coefficient calculation and the logged-loss check
answer separate questions; a recorded return cannot substitute for verifying
which weight entered backpropagation.

\subsection{Action Parsing and Separate Reliability Counters}

The Qwen input contains observation, permitted history, and explicit legal
actions. A generation is parsed to one candidate action and checked for
legality; invalid attempts trigger the allowed retries, and exhaustion
triggers the documented fallback. In the initial inspected failure, three
attempts precede fallback. Ambiguous output must not be silently converted
into a legal candidate. First-pass invalidity, total invalid attempts,
parser failures, retries, fallback executions, and final legality have
separate counters. The executed trajectory uses the final action, whereas
the trace retains the model's proposals and the wrapper's decisions.

\subsection{Evaluation Schedule and Budget Accounting}

For historical seed-wise summaries, S1--S10 denote training seeds
20260800--20260809 in ascending order: S1--S3 form the initial batch and
S4--S10 the added batch.

The initial mapping pairs training seeds 20260800--20260802 with evaluation
seeds 20260810--20260812, respectively. For one checkpoint, three opponents
and two seats each receive ten games, giving 60 games. Thus each arm has
180 games over three checkpoints, and all four arms have 720. The added batch
uses all three evaluation seeds for every checkpoint: $3\times3\times2\times10
=180$ games each, $7\times180=1{,}260$ per arm, and 5,040 in total.
These schedules preserve opponent/seat balance but use different
evaluation-seed coverage. They do not replace common-runtime evaluation of
the same weights.

\begin{table}[ht]
\caption{Training-to-evaluation correspondence. Every evaluation seed uses
three opponents, two seats, and ten episodes per opponent--seat cell.
All four arms share the indicated schedule within each training-seed block.}
\label{tab:seed-evaluation-map}
\centering\small
\begin{tabular}{lllrr}
\toprule
Batch & Training seed & Evaluation seed(s) & Games/checkpoint & Games/arm\\
\midrule
Initial & 20260800 & 20260810 & 60 & \\
 & 20260801 & 20260811 & 60 & \\
 & 20260802 & 20260812 & 60 & 180\\
Added & 20260803--20260809 & 20260810, 20260811, 20260812 & 180 & 1,260\\
\bottomrule
\end{tabular}
\end{table}

The episode pairing key is (training seed, evaluation seed, opponent, seat,
episode index); arm is the matched comparison factor. Distinct episode keys
need not produce distinct paths. Method-level summaries first compute each
within-seed arm contrast using its scheduled games, then average seed
contrasts with equal weights within the batch. Reusing an evaluation seed
does not create another training replication.

\paragraph{Validation splits have a different role.}
Reward-gate development and confirmation states are tracked by canonical
information-state identity and historical exposure, separately from the
training/evaluation seeds above. Leduc v0.1 uses 73 development and 24
confirmation states; its $9/24$ result fails the gate. All 24 states in the
later Leduc v0.2 $20/24$ re-validation had already appeared in historical
development data. Goofspiel v0.1 and v0.2 report $7/29$ and $18/29$ on their
respective confirmation protocols. Appendices~\ref{app:leduc-gate} and
\ref{app:goofspiel-gate} give the state exposure and stopping records.
A new split name or random seed does not establish fresh confirmation
evidence.

\paragraph{Commands and the results they reproduce.}
The supplied summary verifier uses only the Python standard library. It
embeds the exact CSV count snapshots underlying the initial and added seed
tables, checks their SHA-256 identifiers, and prints the seed/evaluation map,
integer contrasts, equal-seed means, tie counts, and two-sided exact sign
tests. It keeps the batches separate:
\begin{verbatim}
python verify_replication_units.py --batch initial
python verify_replication_units.py --batch added
python verify_replication_units.py --batch all
\end{verbatim}
These commands reproduce summary arithmetic, not training runs or trajectory
deduplication. The original initial-batch record-level analysis entry point
is \texttt{analysis/analyze\_formal720\_statistics.py} in the statistical
analysis archive, with the following input contract:
\begin{verbatim}
python analysis/analyze_formal720_statistics.py \
  --bundle-root /path/to/glm-v04-formal-720-delivery-20260829 \
  --audit-trace /path/to/full_cache_audit_trace.jsonl \
  --output-dir analysis/results --bootstrap-replicates 100000
\end{verbatim}
That entry point reads per-arm episode files from each training-seed folder;
its game-record intervals retain the nominal status specified in
Appendix~\ref{app:statistical-diagnostics}. The separate terminal-replay
command and its 12 raw input files are documented in
Appendix~\ref{app:terminal-replay}. The completed fixed-checkpoint common-runtime evaluation and any retraining
require the corresponding executed source, configuration, and checkpoint
identities in Appendix~\ref{app:runtime-identity}; the summary commands do
not stand in for those experiments.

Costs retain separate units:
\begin{equation}
 B=(N_{\rm env},T_{\rm GPU},N_{\rm annotation},
 N_{\rm policy\text{-}infer},N_{\rm judge\text{-}train},C_{\rm API}),
 \label{eq:budget}
\end{equation}
together with token counts and per-inference latency. Offline annotation is
charged once to its reward artifact. The read-only cache provider implies
$N_{\rm judge\text{-}train}=0$; this does not imply zero Qwen inference or GPU
training cost. Episode counts, optimizer steps, and action examples remain
separate budget units. Environment calls during training and evaluation are
reported separately, as are lookup support and nonzero dense exposure.

For completeness, the crossed comparison also admits average system
contrasts $\Delta_{\rm reward}=(\Delta_T+\Delta_Q)/2$ and
$\Delta_{\rm policy}=[(\mu_P-\mu_B)+(\mu_H-\mu_R)]/2$. Their arithmetic is
well-defined, but their interpretation inherits the distinct learners,
data volumes, and execution identities. The main cases use the paired
within-family contrasts and interaction to direct diagnosis.

\subsection{Comparing Batches Without Conflating Sources of Change}

A second batch introduces two possible sources of change: new training
realizations and a changed measurement or execution contract. LocusRL first
records both, then asks which observations a candidate explanation can and
cannot account for. For batch $b$, seed $s$, and evaluation stratum $z$,
\begin{equation}
 \Delta_{Q,b,s}=\sum_z q_b(z)
   \bigl[\widehat\mu_{H,b,s,z}-\widehat\mu_{P,b,s,z}\bigr],
 \label{eq:batch-standardization}
\end{equation}
where $z$ includes opponent, seat, and evaluation schedule, and $q_b$ gives
the empirical weights. More games can change this weighting or realized
paths; they do not create new trained policies. Comparisons should align
measurement support before pooling, while retaining the actual training
seed as the unit of retraining evidence.

This is a diagnostic decomposition of the comparison, not an estimate of
how much each layer caused the batch difference. When source equivalence is
unresolved, the next discriminating test fixes checkpoint weights and varies
only the evaluation runtime. When the declared objective and numerical logs
disagree, source and batch-level reconciliation precede reward-effect claims.
Each unresolved check therefore identifies a next experiment, preserving
the distinction between localization and causal isolation.

\FloatBarrier
\section{Historical Connect Four Diagnostics}
\label{app:historical}

The first run uses only 74 valid V1 cache entries. Reward training coverage is
$269/10{,}881=2.47\%$, and Policy/Hybrid fallback reaches $32.14\%$ and
$73.73\%$. Its arm win rates are $40.56\%$, $44.44\%$, $46.11\%$, and
$41.11\%$; it is retained only as a support-collapse diagnostic.

\begin{table}[ht]
\caption{Second 720-game run using low-token DeepSeek V2. Reward quality fails
the later tactical audit, so these outcomes are historical negative evidence.}
\label{tab:historical-four-arm-results}
\centering
\scriptsize
\setlength{\tabcolsep}{4pt}
\begin{tabular}{lrrrrrrr}
\toprule
Agent & Random & Heur. v1 & Heur. v2 & Overall & Seat 0 & Seat 1 & Fallback\\
\midrule
Base   & $90.00\%$ & $50.00\%$ & $0.00\%$ & $46.67\%$ & $62.22\%$ & $31.11\%$ & $0.00\%$\\
Reward & $95.00\%$ & $50.00\%$ & $0.00\%$ & $48.33\%$ & $63.33\%$ & $33.33\%$ & $0.00\%$\\
Policy & $86.67\%$ & $50.00\%$ & $50.00\%$ & $62.22\%$ & $95.56\%$ & $28.89\%$ & $1.84\%$\\
Hybrid & $86.67\%$ & $50.00\%$ & $33.33\%$ & $56.67\%$ & $84.44\%$ & $28.89\%$ & $4.31\%$\\
\bottomrule
\end{tabular}
\end{table}

For provenance, the pre-revalidation four-category snapshot is retained here
as a legacy diagnostic: its V2 sign agreement was
$3{,}060/9{,}954=30.74\%$, exact score agreement $16.36\%$, mean absolute
error $0.6596$, and $39.76\%$ of entries conflicted with its preregistered
tactical thresholds. In that legacy labeling, 2,945 of 3,712 items called
``obvious errors'' received positive reward. These counts are not the current
five-category reference; the revised source of truth is
Appendix~\ref{app:reference-revalidation}.

\FloatBarrier
\section{Auxiliary Reward-Mechanism Controls}
\label{app:mechanism-controls}

\paragraph{Short deterministic-oracle control.}
An auxiliary five-seed control compares outcome-only learning with a
historical deterministic tactical oracle and its exact sign flip. These
completed runs predate the reference revision in
Appendix~\ref{app:reference-revalidation}. Within each seed, the three
conditions share the recorded initialization hash. Each run performs
five optimizer updates with four game rollouts per update and evaluates 30
games; the configured \texttt{episodes=200} field is not consumed by the
training loop and is therefore not interpreted as 200 completed games.

\begin{table}[ht]
\caption{Short deterministic reward controls. These are mechanism checks,
not GLM v0.4 ablations.}
\label{tab:oracle-controls}
\centering
\small
\begin{tabular}{lrr}
\toprule
Condition & Wins/games & Win rate\\
\midrule
Outcome only & $48/150$ & $32.00\%$\\
Deterministic oracle dense & $56/150$ & $37.33\%$\\
Sign-flipped deterministic oracle & $58/150$ & $38.67\%$\\
\bottomrule
\end{tabular}
\end{table}

The sign-flipped condition does not exhibit the expected degradation. Thus
the control confirms that distinct reward providers, traces, and checkpoints
are wired and archived, but it does not validate reward-direction sensitivity
or explain the formal GLM results. Evaluation seeds also depend on Python's
process-salted \texttt{hash(opponent\_name)} without a recorded
\texttt{PYTHONHASHSEED}; the archived episode records, rather than command-level
trajectory identity, are the reproducible object.

\paragraph{Fixed-data Qwen coefficient sensitivity.}
The recovered rows reproduce the recorded centered returns. They do not
verify the executed weights $w_t$ in Equation~\ref{eq:qwen}, whose identity
requires reconciliation with the loss logs.
For seeds 20260800--20260802, nonzero dense-return counts are 464, 448,
and 463. The ratios $\operatorname{sd}(D_c)/\operatorname{sd}(A_0)$ are
$0.362$, $0.383$, and $0.359$. Negation on these same rows gives
$A_-=A_0-D_c$ and $A_++A_-=2A_0$. Relative to $A_+$, this changes
120, 198, and 116 coefficient signs, with coefficient cosines $0.772$,
$0.748$, and $0.776$. Negating only the dense component does not negate
the entire training signal: the outcome term remains fixed. These are
algebraic checks on serialized examples, not optimizer-visit counts,
gradient cosines, executed optimization weights, trained sign-flip checkpoints, or counterfactual win rates.
They are distinct from the historical tactical-oracle experiment above.

\paragraph{Historical DeepSeek v0.3 cache mechanism run.}
A separate five-seed run applies an earlier DeepSeek v0.3 offline cache for
100 optimizer updates with 20 game rollouts per update and evaluates 150 games
per seed. It wins $367/750=48.93\%$; a same-length deterministic-oracle run
wins $349/750=46.53\%$. Initial parameter hashes match between these two
conditions within seed, and all five v0.3 checkpoint hashes reproduce. Across
training, however, the cache hits only
$34{,}529/175{,}321=19.69\%$ of lookups, and only
$1{,}635/175{,}321=0.93\%$ of all lookups receive a nonzero dense term.
The available outcome-only runs use a much shorter training and evaluation
budget, so this experiment cannot establish that v0.3 reward improves over
outcome-only learning. It is retained only as an offline-cache mechanism run
and is not pooled with the formal GLM v0.4 experiment.

\FloatBarrier
\section{GLM v0.4 Cache Audit and Reproducibility}
\label{app:cache-audit}

\subsection{Frozen Record Schema and Runtime Lookup}

Offline annotation writes an immutable JSONL artifact and manifest. A record
contains the state hash, action, acting player, environment commit, provider,
model, available revision metadata, prompt hash, reward version, raw response,
parsed score, token usage, retries, failure reason, validity flag, and creation
time. The API does not expose a model revision; unavailable revision metadata
is recorded as unavailable. A valid record and its digest attest to stored
content, while tactical labels supply the separate semantic check.

The deployed provider, \rewardprovider, has no API client. Construction
rejects missing or tampered cache artifacts. Lookup uses the exact state hash,
action, and player; an absent or invalid entry sets the dense mask to zero
and is counted separately. There is no interpolation to unseen positions and
no fabricated score after exhausted annotation requests. Terminal dense
terms are masked and the environment outcome is counted exactly once.
Reward and Hybrid share the 9,954-entry artifact, split 4,977/4,977 by acting
seat, with $\lambda=1$.

\subsection{Support Denominators and Full-Cache Audit}

For the initial batch, Reward training has 245,559 hits from 717,409 lookups
(34.23\%), and Hybrid has 2,531 from 5,906 (42.85\%). Evaluation has
1,040/2,602 (39.97\%) and 1,000/2,821 (35.45\%), respectively. These count
lookup events, rather than unique states or necessarily nonzero scores.
Policy/Hybrid fallback fractions instead use agent-action denominators:
20/1,486 (1.35\%) and 122/1,488 (8.20\%). Their 6.85-point difference is an
action-reliability contrast, distinct from the game-level win-rate contrast.

The archived join against the original tactical reference yields consistent
support of 842/2,602 for Reward and 802/2,821 for Hybrid. These are historical
reference-dependent summaries, preserved in
Table~\ref{tab:evaluation-semantic-support}; they have not been recomputed
against the revised labels. The exact hit counts above are label-independent.
Support and fallback depend on states reached by the policy and do not by
themselves decompose a learning effect.

The complete frozen cache has now been re-audited after human trial cases
exposed false defensive labels. The revised reference checks all immediate
opponent replies after a nonwinning candidate and distinguishes unavoidable
next-reply loss from avoidable tactical error. Table~\ref{tab:glm-cache-audit}
reports this revised reference; Appendix~\ref{app:reference-revalidation}
gives the executable definition, witnesses, and exact old-to-new accounting.

\begin{table}[ht]
\caption{Revalidated deterministic audit of all 9,954 frozen GLM v0.4
scores. Forced loss denotes unavoidable loss on the opponent's next reply;
its negative direction evaluates the state, not blame for the candidate.
Neutral agreement uses $|f|\leq0.2$. These are finite-cache counts.}
\label{tab:glm-cache-audit}
\centering
\small
\begin{tabular}{lrrr}
\toprule
Revised category & Entries & Direction agreement & Positive error/loss\\
\midrule
Required block & 255 & $232$ ($90.98\%$) & --\\
Threat creation & 757 & $618$ ($81.64\%$) & --\\
Avoidable tactical error & 3,444 & $2{,}863$ ($83.13\%$) & $556$ ($16.14\%$)\\
Forced next-reply loss & 471 & $416$ ($88.32\%$) & $52$ ($11.04\%$)\\
Neutral & 5,027 & $3{,}813$ ($75.85\%$) & --\\
\midrule
All & 9,954 & $7{,}942$ ($79.79\%$) & --\\
\bottomrule
\end{tabular}
\end{table}

\begin{figure*}[t]
  \centering
  \includegraphics[width=0.897739\linewidth]{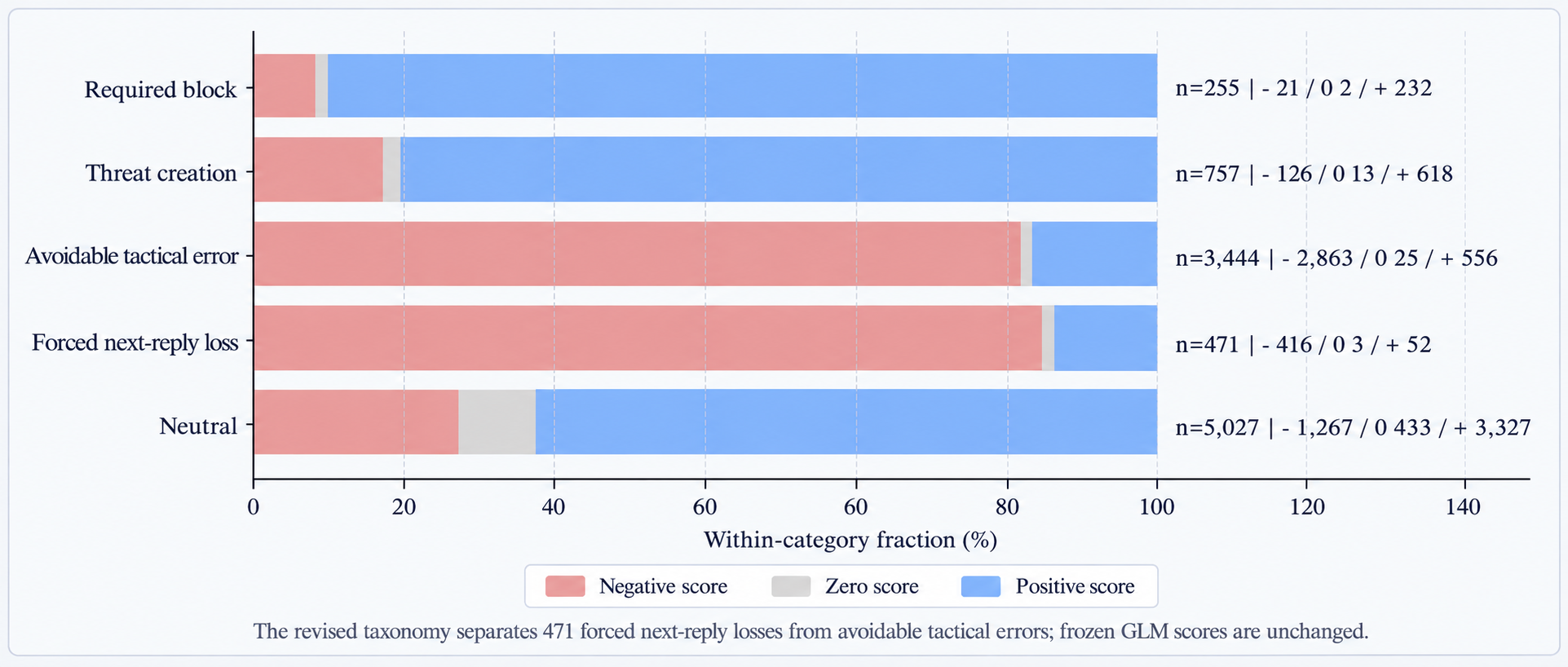}
  \caption{Score-sign composition under the revised reference. Each horizontal
  bar partitions one category into negative, zero, and positive frozen GLM
  scores. The taxonomy separates 471 forced next-reply losses from 3,444
  avoidable tactical errors; no GLM score is recomputed. Exact direction
  agreement is reported in Table~\ref{tab:glm-cache-audit}; neutral agreement
uses $|f|\leq0.2$, not the zero-score share.}
  \label{fig:reward-category-distribution}
\end{figure*}

With the original numeric thresholds applied to the revised categories,
the threshold pass rate is $8{,}567/9{,}954=86.07\%$ (historically
$8{,}498/9{,}954=85.37\%$). Strict JSON is available for 9,752 entries;
202 are recovered by the production parser and no API-failure entries remain.
No chosen action is an immediate win, limiting category coverage.

Independent verification now succeeds for $9{,}954/9{,}954$ cache entries and
all $9{,}954/9{,}954$ recorded entry digests. The independently recomputed
aggregate manifest hash equals the declared value:
\begin{center}
\small\texttt{34d56037688c81550a2734a193438e1dd}\linebreak
\texttt{735211f2c0f6499e062f1a4fae062fe}.
\end{center}
The frozen rubric hash also matches. Board-content joining succeeds for every
entry.

The canonical v0.4 builder appends the must-block rubric to the v0.3
state/action user prompt. Recomputing this dynamic prompt reproduces
$9{,}954/9{,}954$ stored prompt hashes, with no source-reference or
state--action--player mismatches. The repository's public canonical-pool file
remains bytewise distinct from the original absolute-path source named in the
manifest, although all 9,954 records join by content; this is a source-path
normalization issue rather than a cache-integrity failure.

\paragraph{Evaluation archive and training-evidence boundary.}
The formal delivery archive has SHA-256:
\begin{center}
\small\texttt{505c3f2f2c8259b9ebf924ffb74f4abd}\linebreak
\texttt{2019067ffbe76ef026a7ceda1b18eeee}.
\end{center}
All 60 files listed in its package-relative checksum manifest verify,
including 720 per-game
evaluation records, 12 arm/seed training configurations, 12 training
summaries, and the available Reward and evaluation traces. The original
lightweight archive excludes model weights and omitted the per-step
\texttt{train\_examples.jsonl} and Hybrid training traces. A later recovery
from the original run roots locates 12 canonical training-example files and
one Hybrid training trace per seed. Their 15 individual SHA-256 values are
recorded, together with recovery-archive SHA-256
\texttt{ab2710b92cb283164b0209492f645dee780d9f340f821b5faa89a50f87f8284f}.
No training or evaluation was rerun and no substitute rows were introduced.
A public reproducibility package at commit \texttt{f04b802} now contains that
recovery archive alongside the 720 per-game records, checkpoint identities,
environment and model commits, run commands, and cache audit. The archive has
13 physical training-example paths because one seed-20260801 Base source copy
is byte-identical to its canonical counterpart; the evidence map selects 12
unique seed--arm files and records the duplicate explicitly. Independent
inspection reproduces the four arm totals and validates the recovery archive's
internal digests. The final delivery reports that, from a clean checkout,
\texttt{python -B verify\_bundle.py} passes; the 46-entry top-level SHA-256 manifest excludes
\texttt{\_\_pycache\_\_} and \texttt{*.pyc}, and an equivalent per-file
SHA-256 check passes on Windows. The final compressed package is 6,113,201
bytes with SHA-256:
\begin{center}
\small\texttt{88fe78337d6ac7b55c078d03a8bd6e5b}\linebreak
\texttt{8e2709169af5f046e868c2dcc2c9e732}.
\end{center}

\FloatBarrier
\section{Training and Evaluation Runtime Identities}
\label{app:runtime-identity}

\begin{table}[t]
\caption{Runtime and seed sensitivity of the recorded contrasts. Qwen columns
give Policy/Hybrid labels; effects are percentage points on 60 games per arm
and seed. A is \texttt{fe1c288}; B is \texttt{f458d5d}. Labels are provenance
records, not a measured runtime treatment.}
\label{tab:main-attribution}
\centering\small
\setlength{\tabcolsep}{5pt}
\begin{tabular}{lccrrr}
\toprule
Training seed & Qwen train & Qwen eval & Reward--Base & Hybrid--Policy & Interaction\\
\midrule
20260800 & A/A & A/B & $+1.67$ & $-16.67$ & $-18.33$\\
20260801 & B/B & B/B & $+1.67$ & $0.00$ & $-1.67$\\
20260802 & B/B & B/B & $+1.67$ & $0.00$ & $-1.67$\\
\bottomrule
\end{tabular}
\end{table}

\paragraph{Arm-by-phase reconstruction.}
We read twelve training and twelve evaluation summaries from the archived
\path{glm-v04-formal-720-delivery-20260829.tgz}. The files are stored separately
by seed, phase, and arm. Table~\ref{tab:runtime-identity} uses their
\texttt{env\_commit\_hash} field rather than a seed-level source-root summary.
All 24 summaries record the same \texttt{commit\_hash}, A. Within each seed,
Base/Reward share the recorded initialization parameter hash, as do
Policy/Hybrid; this pairing survives the phase-level audit.

\begin{table}[ht]
\caption{Recorded training/evaluation environment labels. Each cell is
training / evaluation. A and B are defined below. The highlighted cell
identifies the within-pair evaluation discrepancy; labels alone do not
establish an executed code difference.}
\label{tab:runtime-identity}
\centering\small
\begin{tabular}{lcccc}
\toprule
Training seed & Base & Reward & Policy & Hybrid\\
\midrule
20260800 & A / A & A / A & A / A & \textbf{A / B}\\
20260801 & B / B & B / B & B / B & B / B\\
20260802 & B / B & B / B & B / B & B / B\\
\bottomrule
\end{tabular}
\end{table}

\noindent\begin{minipage}{\linewidth}
A: \texttt{fe1c28817de070ee97088d25490a00206f205ed0}.\\
B: \texttt{f458d5da63e0c4ef96c310fb7d0e115956081753}.
\end{minipage}

For seed 20260800, the checkpoint hash prefixes are \texttt{85f83fea6ac5}
(Policy) and \texttt{2fc89d6c4c33} (Hybrid).
Both checkpoints were trained under recorded environment A, but the Hybrid
fixed evaluation records B. A seed-level statement that the entire
\texttt{final2} block uses A misses this evaluation-specific discrepancy.

\paragraph{Metadata does not certify source equivalence.}
In the delivered integration-branch source, the provenance module obtains
\texttt{ENV\_COMMIT\_HASH} from \texttt{LOCUSRL\_ENV\_COMMIT\_HASH}, with B as
its default. Its Git metadata helper also accepts environment-variable
overrides. These mechanisms mean that a recorded label can differ without
proving that game rules changed. Conversely, they do not establish that the
historical discrepancy was only a label error: the delivered source is not
proof of which files were imported during every historical process.

The archived phase labels remain useful provenance evidence, but they do not
by themselves prove a causal runtime effect. This ambiguity is now separated
from checkpoint behavior by a completed fixed-checkpoint common-runtime
re-evaluation.

\paragraph{Completed common-runtime re-evaluation.}
All ten historical training seeds are evaluated under one frozen runtime and
one common opponent/seat schedule, without retraining. Each seed contributes
20 games per arm (random\_v1 and heuristic\_v1, balanced by seat), giving
200 games per arm and 800 games overall. The initial three-seed subset records
Base/Reward/Policy/Hybrid wins of 37/37/34/29; the added seven-seed subset
records 96/96/97/96; totals are 133/133/131/125. Re-evaluating the same
checkpoint does not add a training replicate. Because the schedule differs
from the historical evaluation schedules, the old-to-new change is not
identified as a pure runtime causal effect; it is a fixed-weight result under
one shared evaluation contract. The initial historical Policy/Hybrid training-instance logs are still
needed to establish completed updates and actual example consumption;
the added batch already has 1,400 numerical update records. Full historical
objective/source correspondence is a separate recovery question.

\FloatBarrier
\section{Paired Statistical Diagnostics and Sensitivity}
\label{app:statistical-diagnostics}

\paragraph{Units of evidence.}
The initial batch analyzed in this appendix contains 720 evaluation games and twelve trained
checkpoints: four arms under each of three paired training-seed blocks. The
added batch has 5,040 games and 28 checkpoints under seven blocks. All arms,
evaluation seeds, opponents, seats, and episode records belonging to a
training seed stay together for the method comparison. The batch effect is
the equal-weight mean of its seed contrasts. Counts of checkpoints or games
do not increase the number of independently seeded training blocks.

\paragraph{Effective independent sample size.}
An effective sample size depends on an estimand and a dependence model;
neither record count nor trajectory deduplication identifies a universal
value. We therefore disclose the actual units: the ten initial Qwen
discordances have one paired-path signature and one contributing training
seed; the 20 Hybrid repair certificates share one first-repair board and two
checkpoints, with zero prospective repaired-policy rollouts. The 72 joint
trajectory classes in Appendix~\ref{app:terminal-replay} describe behavioral
coverage, not 72 demonstrated independent draws. The certificates are exact
checks on the specified recorded prefixes, not Bernoulli repair trials; we
do not attach a binomial success interval to their $20/20$ coverage.

\paragraph{Inferential model and nominal record-level calculations.}
Our method-level test conditions on the nonzero paired training-seed
contrasts and assumes that their signs are independent and equiprobable
under the null of no directional tendency. Runtime or schedule heterogeneity
remains part of the interpretation; the test does not identify a causal
reward effect. Exact McNemar tests and the stratified bootstrap below are
retained as nominal game-record calculations. Their game-level stochastic
or exchangeable sampling model is not established by repeated paths. They
provide neither calibrated within-checkpoint uncertainty under arbitrary
repetition nor independent method replications. Multiplicity adjustment
cannot repair that dependence assumption.

\paragraph{Resolution depends on nonzero seed contrasts.}
For a two-sided exact sign test that omits ties, $k$ nonzero contrasts have
minimum attainable $p=2^{1-k}$. Table~\ref{tab:seed-sign-resolution} derives
$k$ from the unrounded win counts in each batch. In particular, the added
Qwen comparison has four nonzero signs, not seven; its observed $p=1$.
The six nonzero added interaction signs have minimum $.03125$, but their
observed split gives $p=.6875$. These diagnostic calculations do not resolve
runtime or optimizer identity, and the two batches remain separate.
Writing $n_+$ and $n_-$ for the positive and negative counts, respectively,
the exact calculation is
\[
 k=n_++n_-,\qquad
 p=\min\!\left\{1,\;2^{1-k}
       \sum_{j=0}^{\min(n_+,n_-)}\binom{k}{j}\right\},
 \quad k>0.
\]
Exact zero contrasts are omitted; with no nonzero signs we report $p=1$
by convention. Here $k$ is the number of informative signs, not an estimated
effective sample size. The supplied verifier implements this expression
with integer arithmetic; Appendix~\ref{app:protocol} maps commands to the
two source snapshots.

\begin{table}[ht]
\caption{Exact training-seed sign-test resolution by batch. Ties are omitted
from $k$; signs are computed from integer win counts, not rounded effects.}
\label{tab:seed-sign-resolution}
\centering\small
\begin{tabular}{llrrrrr}
\toprule
Batch & Contrast & Positive & Negative & Ties & Minimum $p$ & Observed $p$\\
\midrule
Initial (3) & Reward--Base & 3 & 0 & 0 & $.25$ & $.25$\\
Initial (3) & Hybrid--Policy & 0 & 1 & 2 & $1$ & $1$\\
Initial (3) & Interaction & 0 & 3 & 0 & $.25$ & $.25$\\
Added (7) & Reward--Base & 2 & 3 & 2 & $.0625$ & $1$\\
Added (7) & Hybrid--Policy & 2 & 2 & 3 & $.125$ & $1$\\
Added (7) & Interaction & 4 & 2 & 1 & $.03125$ & $.6875$\\
\bottomrule
\end{tabular}
\end{table}

\begin{table}[ht]
\caption{Nominal paired calculations for the formal 720-game evaluation.
The previously computed intervals resample matched game records within seed
blocks; repeated paths limit their inferential interpretation even with
checkpoints fixed. Holm adjustment covers the
two aggregate within-backbone McNemar tests (Reward--Base and
Hybrid--Policy), with no other tests included.}
\label{tab:paired-inference}
\centering
\scriptsize
\setlength{\tabcolsep}{3pt}
\begin{tabularx}{\linewidth}{lrrrrY}
\toprule
Contrast & Effect & Nominal 95\% interval & Discordant & Nominal exact $p$ & Evidence boundary\\
\midrule
Reward--Base & $+1.67$ pp & $[-1.67,+5.00]$ & $6$ vs. $3$ & $.508$ & Holm $p=.508$; seed-sign $p=.25$\\
Hybrid--Policy & $-5.56$ pp & $[-8.89,-2.78]$ & $0$ vs. $10$ & $.00195$ & Holm $p=.00391$; seed-sign $p=1$; one paired path in one seed\\
Interaction & $-7.22$ pp & $[-11.67,-2.78]$ & -- & -- & Seed-sign $p=.25$; not a game-level binary contrast\\
\bottomrule
\end{tabularx}
\end{table}

\begin{figure*}[t]
  \centering
  \includegraphics[width=0.98\linewidth]{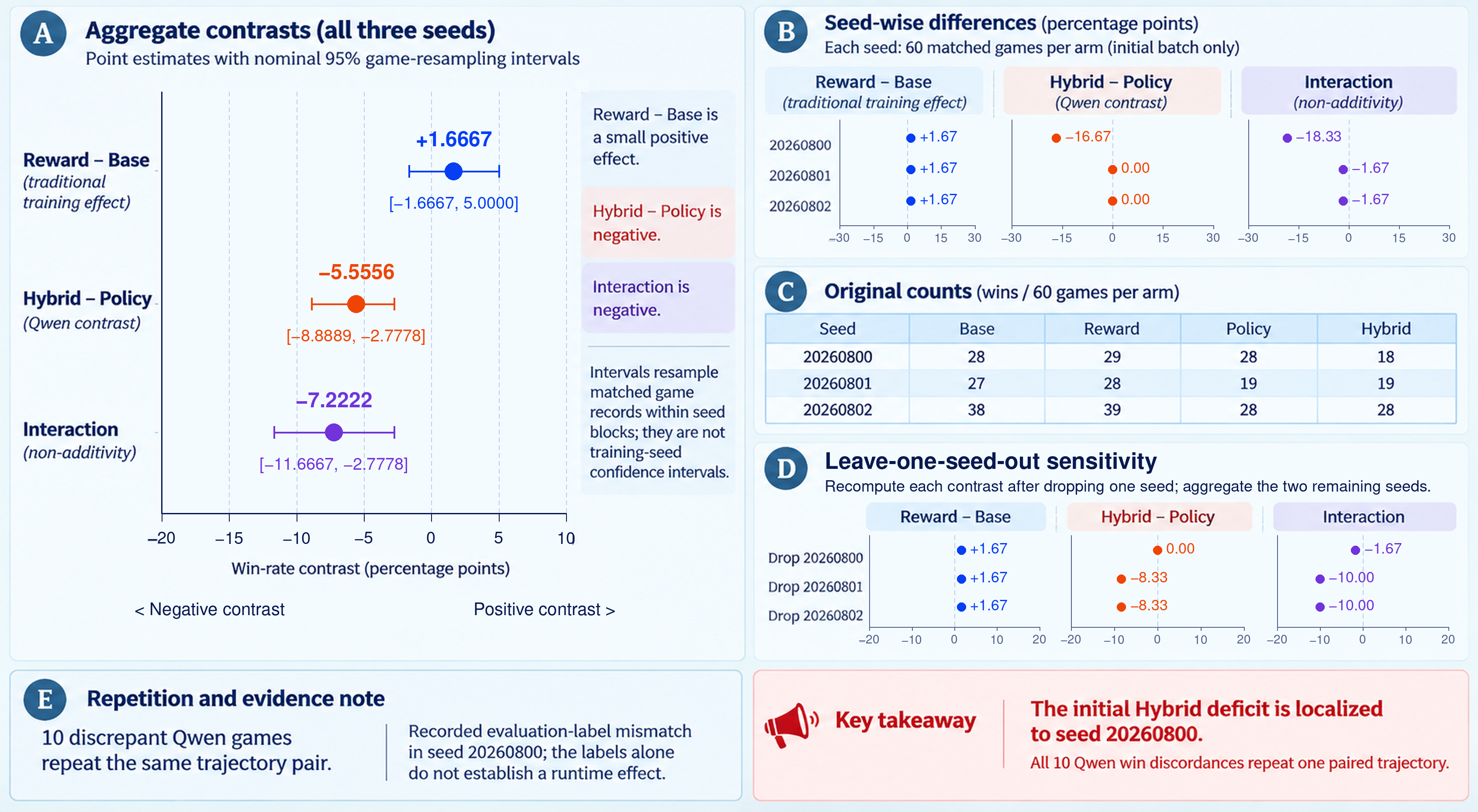}
  \caption{Localization of the initial three-seed contrasts. Panel A gives recorded aggregate differences with nominal 95\% game-resampling intervals; panels B and C give seed differences and win counts, and Panel D gives leave-one-seed-out estimates. Only the aggregate intervals have a documented resampling interpretation; the local per-seed bars are not used for inference. All ten Qwen win discordances repeat one paired path in seed 20260800. The aggregate intervals are not training-seed confidence intervals.}
  \label{fig:paired-effect-localization}
\end{figure*}

\paragraph{Seed localization and leave-one-out sensitivity.}
The traditional reward contrast is $+1.67$ points in each seed. The Qwen
contrast is $-16.67$, $0.00$, and $0.00$ points, producing interactions of
$-18.33$, $-1.67$, and $-1.67$ points. Dropping seed 20260800 leaves
Reward--Base at $+1.67$ points, Hybrid--Policy at $0.00$, and the interaction
at $-1.67$. Dropping either of the other seeds instead yields a
Hybrid--Policy contrast of $-8.33$ points and an interaction of $-10.00$.
This sensitivity establishes localization, not the direction of a causal
reward effect.

\paragraph{Semantic support in observed evaluation states.}
Exact cache coverage counts any hit. The archived effective-support metric
counts hits agreeing with the original tactical reference; inconsistent
exposure counts its disagreeing hits over all lookups.
Table~\ref{tab:evaluation-semantic-support} preserves the earlier join by
state hash, action, and player. Its semantic columns require a fresh join of
the original evaluation lookup traces to the revised reference in
Appendix~\ref{app:reference-revalidation}. They are not updated by substituting
the full-cache agreement rate, because the policies visit a different
distribution. Exact support and the seed outcome tests are unchanged.

\begin{table}[ht]
\caption{Archived evaluation support under the original tactical reference.
Exact support is label-independent; the remaining columns retain the old
reference and are not revised-reference estimates. All columns describe
evaluation visits rather than training exposure.}
\label{tab:evaluation-semantic-support}
\centering
\scriptsize
\setlength{\tabcolsep}{4pt}
\begin{tabular}{lrrrr}
\toprule
Arm & Exact support & Old consistent support & Old inconsistent exposure & Old hit agreement\\
\midrule
Reward & $1040/2602$ ($39.97\%$) & $842/2602$ ($32.36\%$) & $198/2602$ ($7.61\%$) & $842/1040$ ($80.96\%$)\\
Hybrid & $1000/2821$ ($35.45\%$) & $802/2821$ ($28.43\%$) & $198/2821$ ($7.02\%$) & $802/1000$ ($80.20\%$)\\
\bottomrule
\end{tabular}
\end{table}

\paragraph{Why higher-complexity models are not reported.}
Within the initial batch, a mixed-effects interaction model cannot separately estimate a runtime effect
from the Qwen reward effect because the within-pair evaluation-label mismatch is
perfectly localized to the only seed with a nonzero Qwen contrast, and there
are only three training-seed clusters. Similarly, arm-level correlations
between support and wins would use at most six reward-enabled arm--seed points
and would relate post-training evaluation exposure to performance; they would
not identify training support as a mechanism. A naive Beta--Binomial analysis
would provide smooth posterior probabilities but retain the same independence
error if games were treated as method replications. We therefore report exact
nominal paired game tests and bootstrap summaries, seed-level sign tests, and
leave-one-seed-out sensitivity rather than a more elaborate but unidentified
model.

\FloatBarrier
\section{GLM v0.4 Seed and Coverage Breakdown}
\label{app:glm-breakdown}

\begin{table}[t]
\caption{GLM v0.4 Connect Four evaluation. Each arm has 60 games per opponent
and 180 overall. Fallback is pooled over agent actions; all final actions are
legal.}
\label{tab:glm-four-arm-primary}
\centering
\scriptsize
\setlength{\tabcolsep}{4pt}
\begin{tabular}{lrrrrrrr}
\toprule
Agent & Random & Heur. v1 & Heur. v2 & Overall & Seat 0 & Seat 1 & Fallback\\
\midrule
Base   & $88.33\%$ & $50.00\%$ & $16.67\%$ & $51.67\%$ & $72.22\%$ & $31.11\%$ & $0.00\%$\\
Reward & $93.33\%$ & $50.00\%$ & $16.67\%$ & $53.33\%$ & $75.56\%$ & $31.11\%$ & $0.00\%$\\
Policy & $91.67\%$ & $33.33\%$ & $0.00\%$ & $41.67\%$ & $52.22\%$ & $31.11\%$ & $1.35\%$\\
Hybrid & $91.67\%$ & $16.67\%$ & $0.00\%$ & $36.11\%$ & $41.11\%$ & $31.11\%$ & $8.20\%$\\
\bottomrule
\end{tabular}
\end{table}

\begin{figure*}[t]
  \centering
  \includegraphics[width=0.98\linewidth]{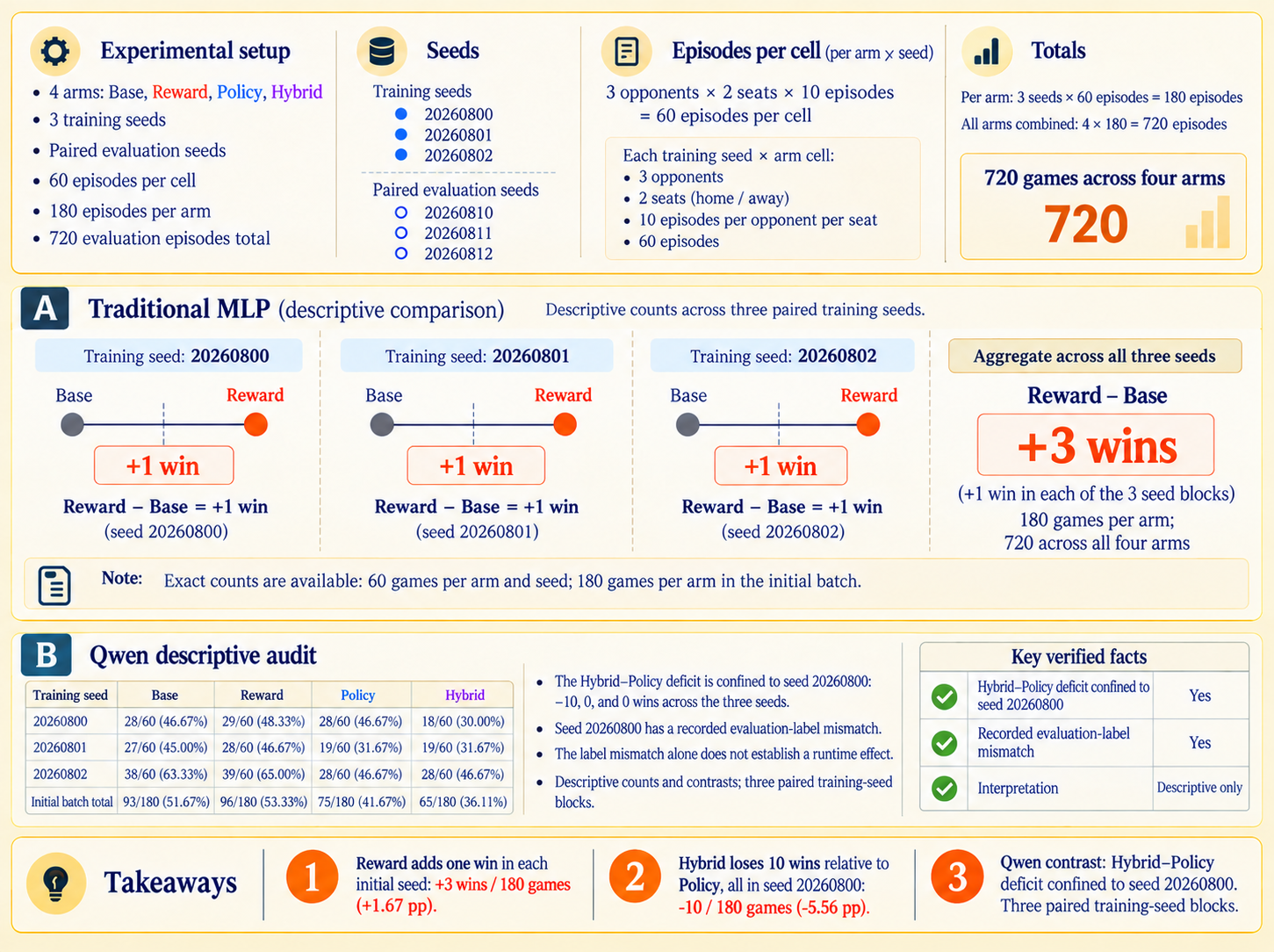}
  \caption{Seed-structured descriptive summary of the formal 720-game
  evaluation. Reward adds one win over Base in each of the three paired
  training-seed blocks. The aggregate Policy--Hybrid difference is confined
  to seed 20260800, whose Qwen pair records different evaluation environment
  labels. The other two blocks have matched labels. The recorded evaluation-label mismatch is marked in the figure; executed source equivalence remains unverified. Nominal matched-game calculations are reported in
  Appendix~\ref{app:statistical-diagnostics}; method-level generalization
  remains bounded by the three training seeds and runtime mismatch.
  Exact aggregate rates are reported in Table~\ref{tab:glm-four-arm-primary},
  and the per-seed counts are reproduced in
  Appendix~\ref{app:glm-breakdown}.}
  \label{fig:formal720-seeds}
\end{figure*}

Across the three 60-game training/evaluation blocks, Base wins $(28,27,38)$,
Reward $(29,28,39)$, Policy $(28,19,28)$, and Hybrid $(18,19,28)$ games.
Thus Reward adds one win in every block, whereas the aggregate Hybrid deficit
comes entirely from the first block.

\begin{figure*}[t]
  \centering
  \includegraphics[width=0.98\linewidth]{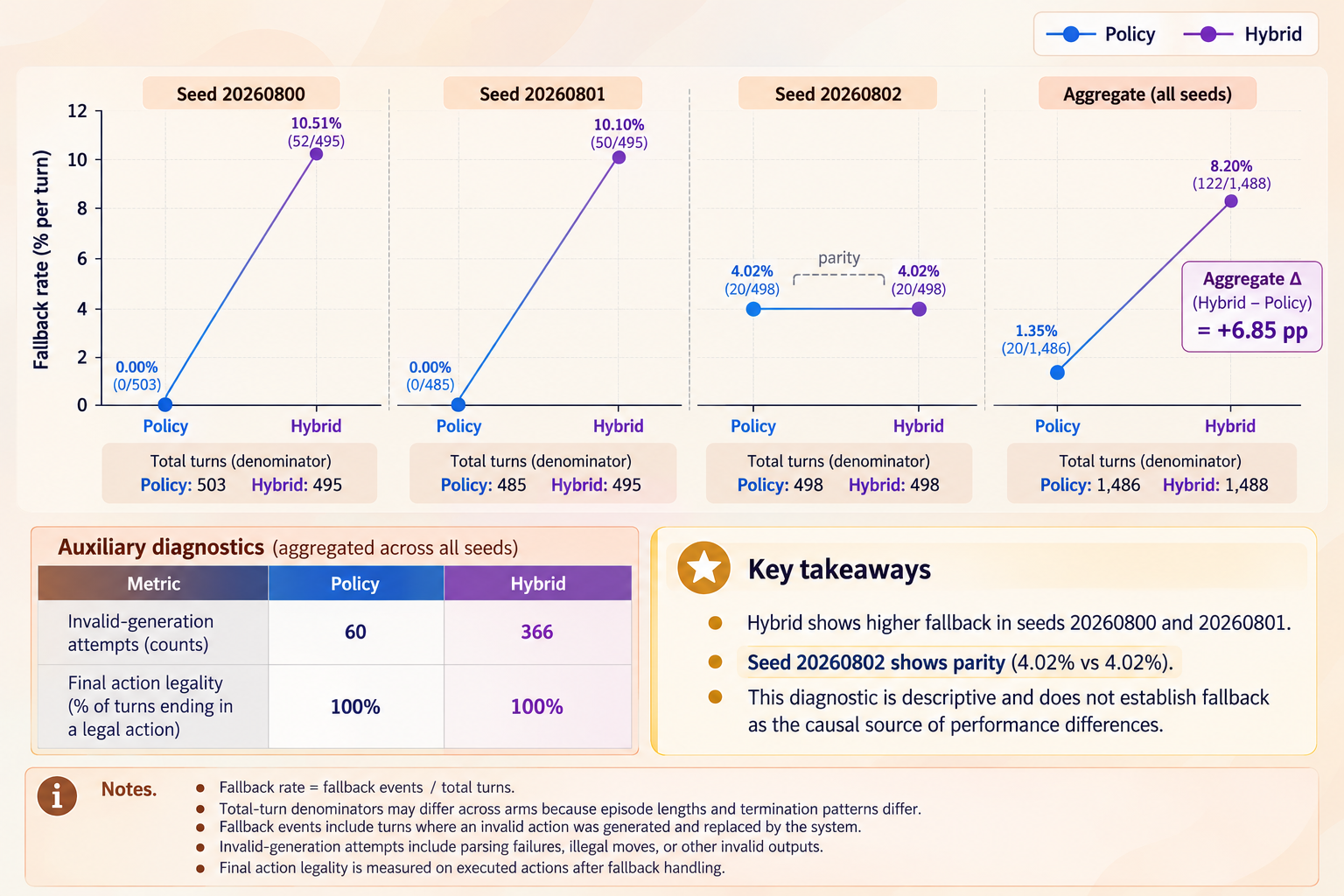}
  \caption{Seed-level action-reliability diagnostic for the two Qwen-policy
  arms. Hybrid has higher fallback in seeds 20260800 and 20260801, while both
  arms are at $20/498=4.02\%$ in seed 20260802. Aggregated across seeds,
  fallback is $20/1{,}486=1.35\%$ for Policy and
  $122/1{,}488=8.20\%$ for Hybrid, a $6.85$-point difference.
  The first two Policy denominators are 503 and 485, both with zero fallback.
  The detailed panel includes the archived episode metrics and auxiliary execution diagnostics.
  Invalid-generation attempts are 60 and 366, respectively, while all
  executed actions are legal after fallback handling. This diagnostic is
  descriptive and does not identify fallback as the causal source of outcome
  differences.}
  \label{fig:fallback-by-seed}
\end{figure*}

Exact training and evaluation support counts for both reward-enabled arms are
shown in Figure~\ref{fig:cache-fidelity-coverage}. An older top-level
\texttt{resume3} aggregate reports
$984/3{,}024=32.54\%$ for Hybrid evaluation. It is retained only for
provenance: the formal conclusion uses the three verified per-seed traces and
their SHA-256 values, which yield $1{,}000/2{,}821=35.45\%$.

\FloatBarrier
\section{Leduc Prospective Reward-Validity Gate}
\label{app:leduc-gate}

The Leduc attempt uses the project's frozen simplified one-raise protocol and
an environment-specific reward prompt. Prompt and cache-key construction
exclude the opponent's private card, the remaining deck, chance/deal seeds,
terminal returns, and reference labels. Deterministic full-tree CFR enumerates
all 120 ordered physical deals and retains 97 reachable information states
whose reference action-value margin exceeds $0.02$. The development and
confirmation partitions contain 73 and 24 information-state pairs,
respectively, with zero information-state overlap.

\begin{table}[ht]
\caption{Leduc reward calibration and the preregistered fail-closed decision.
Pairwise accuracy always uses all registered pairs as its denominator.}
\label{tab:leduc-validity-gate}
\centering
\footnotesize
\setlength{\tabcolsep}{4pt}
\begin{tabularx}{\linewidth}{lrrrrY}
\toprule
Stage & Pairs & Correct & Failures & Leakage & Decision\\
\midrule
Dev v0.1 & 73 & 36 ($49.3\%$) & $0/146$ & 0 & Use single registered revision\\
Dev v0.2 & 73 & 37 ($50.7\%$) & $0/146$ & 0 & Freeze prompt and open confirmation\\
Confirmation & 24 & 9 ($37.5\%$) & $1/48$ ($2.08\%$) & 0 & Gate L1 FAIL; STOP\\
\bottomrule
\end{tabularx}
\end{table}

The confirmation threshold is pairwise accuracy $\geq75\%$, parser/API
failure rate $\leq5\%$, and exactly zero leakage. The pair containing the
single failed action response remains in the preregistered 24-pair denominator;
using only the 23 pairs with two valid responses would change the registered
estimand. The result is evaluated against the frozen threshold, not described
as significantly below a nominal $50\%$ random baseline.

The audit records upstream Gate L0 commit \texttt{3b7824d1e8cb} and delivery
commit \texttt{0f918484c45b}. The frozen-prompt, confirmation-trace,
confirmation-report, and formal-stop SHA-256 prefixes are, respectively,
\texttt{7ef46eb7345f}, \texttt{1978a7f70d4d},
\texttt{c5481d267cb6}, and \texttt{35599c9b3531}; the evidence bundle retains
the complete digests.
The stop record sets the formal cache and its hash to null. No reward-cache
training follows this gate. The separate environment-terminal-reward
Base/Policy experiment in Appendix~\ref{app:leduc-policy} is complete and
does not rely on release of this candidate reward.

A later environment-owner adapter commit, \texttt{0e09e9742f9d}, only swaps
two independent reset assignments. A 45-test compatibility recheck passes and
does not reopen confirmation or change the frozen stop decision.

If a later Leduc-specific reward passes a newly frozen confirmation gate, the
same crossed-comparison and realization audits can be applied. This is an
extension beyond the evidence required here; the present result already defines
the protocol stop boundary for the inspected candidate.

\paragraph{Subsequent v0.2 re-validation and independence correction.}
A later exploratory pairwise-ranking revision achieved $57/73$ on its
development split and $20/24$ on a renamed confirmation split. The audit
at source snapshot \texttt{5740637} established that all 24 latter states
had already been opened as v0.1 development data and that earlier error
reports informed the prompt redesign. Internal v0.2 dev/confirmation
overlap is zero, but none of the 24 evaluation states is new relative to the
opened v0.1 pool. Consequently, $20/24$ is re-validation, not an independent
Gate L1 pass. A 194-entry cache had already been generated before this
independence correction; it is an exploratory artifact and formal four-arm
training remains unauthorized. This retrospective correction is distinct
from the original prospective v0.1 stop. It illustrates that a split's
validity depends on its exposure history, not only its current filename or
disjointness from the current development split. The original $9/24$ result
and its stop record remain frozen.

\FloatBarrier
\section{Goofspiel Prospective Reward-Validity Gate}
\label{app:goofspiel-gate}

The initial Goofspiel attempt uses a task-specific reward prompt for the repository's
deck-size-five simultaneous-bid protocol. The adapter exposes both public
remaining bid-card sets but hides the first player's pending bid from the
second player. Prompt construction additionally excludes the realized future
prize order, chance seed, terminal return, and all reference labels and values.
An exact stochastic zero-sum dynamic program with matrix-game solves generated
2,948 candidate strategic states. Deterministic stratified selection retained
120 states with a reference action-value margin of at least $0.15$, split into
91 development and 29 confirmation pairs with zero information-state and
strategic-state overlap. Gate L0 replays all $120/120$ selected states and
recomputes all $240/240$ action values without mismatch.

\begin{table}[ht]
\caption{Goofspiel reward calibration across two reward versions. Pairwise
accuracy uses every registered pair; failures count action records. The
versions use disjoint states and different scoring protocols.}
\label{tab:goofspiel-validity-gate}
\centering
\footnotesize
\setlength{\tabcolsep}{4pt}
\begin{tabularx}{\linewidth}{lrrrrY}
\toprule
Stage & Pairs & Correct & Failures & Leakage & Decision\\
\midrule
\multicolumn{6}{l}{\textit{Reward v0.1: separate action-scoring calls}}\\
Dev v0.1 & 91 & 32 ($35.2\%$) & $7/182$ ($3.85\%$) & 0 & Use single registered revision\\
Dev v0.2 & 91 & 21 ($23.1\%$) & $0/182$ & 0 & Freeze prompt and open confirmation\\
Confirmation & 29 & 7 ($24.1\%$) & $0/58$ & 0 & Gate L1 FAIL; STOP\\
\midrule
\multicolumn{6}{l}{\textit{Reward v0.2: one pairwise-ranking call per state}}\\
Dev v0.1 & 91 & 52 ($57.1\%$) & $12/182$ ($6.59\%$) & 0 & Use single registered revision\\
Dev v0.2 & 91 & 67 ($73.6\%$) & $4/182$ ($2.20\%$) & 0 & Freeze prompt and open confirmation\\
Confirmation & 29 & 18 ($62.1\%$) & $2/58$ ($3.45\%$) & 0 & Gate L1 FAIL; STOP\\
\bottomrule
\end{tabularx}
\end{table}

In reward v0.1, the sole prompt revision replaced free-text explanations with a strict
three-valued JSON score. It removed seven truncation failures, but the frozen
development run contained 65 tied pairs and did not recover direction
agreement. Confirmation then failed the same registered criteria used for the
handoff decision: pairwise accuracy at least $75\%$, parser/API failure rate at
most $5\%$, and exactly zero leakage. The result is evaluated against that
threshold; it is not used as a test against a nominal random baseline.

The delivery commit is \texttt{317432b3cf0d}. SHA-256 prefixes identify the
dataset manifest (\texttt{1e57a13e4f42}), frozen prompt (\texttt{e39a159cb51f}),
confirmation trace (\texttt{ce8dae24c71e}), confirmation report
(\texttt{78b5b3de2c99}), formal stop (\texttt{01f25b8a4ce2}), and full audit
(\texttt{ccb26bb21ef8}). The 422 recorded real-API
action calls total 239,660 tokens and \$0.081630. A separate DNS-blocked attempt
made zero calls, used zero tokens, and is excluded from both fidelity and cost.

The formal builder raised the registered fail-closed error before writing an
offline cache; the cache and its SHA-256 are therefore null. No Simplifier
handoff, four-arm checkpoint, or Goofspiel policy evaluation exists. A later Goofspiel-specific reward that passes a newly frozen gate could be
carried into the same realization audits. That extension is not required for
the present claim: the inspected evidence closes the gate-level transfer test
while leaving cross-environment learning effects untested.

\paragraph{Reward v0.2: historically disjoint states, another failed gate.}
The later delivery excludes the opened v0.1 union of 120 strategic states
from the same 2,948-state candidate pool. It selects 120 new states from the
remaining 2,828 candidates and splits them into 91 development and 29
confirmation pairs. Recomputing the strategic-state intersections confirms
zero overlap between the new splits and between either new split and either
old split. This avoids the specific exposure reuse found in Leduc v0.2;
it does not establish independence from all possible development choices.

The redesigned prompt requests a preferred bid and a positive magnitude,
assigning opposite signed scores to the two registered actions. Thinking is
disabled. The sole development revision adds a heuristic about bidding on
the unique largest remaining prize. Table~\ref{tab:goofspiel-validity-gate}
reports both development runs and the final confirmation result. One
unparseable response invalidates both action records for one confirmation pair; retaining
all 29 pairs gives $18/29$, below the unchanged $75\%$ threshold. Across the
three runs, 211 pairwise API calls produce 422 action records, consume
120,289 tokens, and cost a recorded \$0.027042. These units differ from the
422 separate action calls in reward v0.1. No formal cache or four-arm result
is added by the redesign.

\paragraph{Artifact checks and chronology boundary.}
The supplied v0.2 delivery is identified as \texttt{41d77f3}. All eleven
listed file hashes match the supplied bytes. Reconstructing the final prompts
and cache keys from the sealed states reproduces their hashes; reparsing the
raw development-revision and confirmation responses reproduces the scores
and failure counts. The cache builder rejects the archived failed gate.
The dataset-manifest, prompt-freeze, confirmation-trace, and overlap-audit
SHA-256 prefixes are, respectively, \texttt{19646a0e87c2},
\texttt{01cd3a13ca3e}, \texttt{c84b0538b6f0}, and \texttt{7648ef36359d}.

The delivery records state selection before the first v0.2 API call and
unchanged state identities during the prompt revision. However, the retained
dataset manifest was rewritten at 15:48:20 UTC on September 4, after the first
development report at 15:46:51; no original pre-call seal is supplied.
The final prompt-freeze record at 15:50:13 precedes the confirmation report
at 15:50:49, and the archived hashes are consistent with that final prompt.
Thus the files verify historical state disjointness and the final scoring
artifact, while the stronger pre-first-call sealing claim relies on the
delivery account. The surviving source contains only the revised prompt, so
the listed rerun commands also do not reconstruct the first development
prompt. Neither limitation changes the failed-gate decision.

\paragraph{Verified v0.3 development stop.}
The September 6 update reports a 90/30 development/confirmation split drawn
after excluding the 240 previously opened states (2,708 candidates remain),
with accuracy/failure/leakage thresholds of $75\%$, $5\%$, and zero,
and at most one prompt revision. It reports an
original seal snapshot retained through revision. Removing the unique-maximum
heuristic and the free-text reason field gives $58/90=64.4\%$ development
accuracy; the sole revision, intended to discourage default low bidding,
gives $38/90=42.2\%$. Both runs report zero failed action records out of 180
and zero leakage. The delivery attributes the remaining errors to opposite
low-bid and high-bid biases. Since neither candidate meets the target,
confirmation was not opened and no formal cache was released. Subsequent
inspection of the supplied development traces reproduces both 90-pair
results and verifies all 23 listed file hashes. The sealed confirmation file
was hashed without inspecting its examples. This is a verified development
stop, separate from the earlier confirmation failures.

The same update reports recovery of the reward-v0.2 first prompt from a
session draft and 182 development traces, while the original confirmation
file bytes and first-prompt freeze manifest remain unavailable. It also
attributes the v0.2 failed response to truncated JSON rather than empty
content. Those recovery artifacts were not part of the independently inspected bundle.
Accordingly, the verified claim remains one unparseable confirmation pair and
a failed gate; the stronger historical sealing claim is not upgraded. This is
an evidence boundary for that historical delivery, not an outstanding training
experiment.

\paragraph{Reference target sensitivity on exposed v0.3 states.}
A separate audit of the 90 opened development pairs varies the opponent's
equilibrium mixture while retaining the original continuation values.
If $M$ is the actor's payoff matrix and $v$ the game value, it minimizes and
maximizes $(M_a-M_b)q$ subject to $q\geq0$, $\mathbf{1}^{\top}q=1$, and
$Mq\leq v\mathbf{1}$. At tolerance $10^{-7}$, 47 pairs retain a strict
direction, 42 admit a tie, and one admits reversal. The original values
recompute to maximum absolute error $5.55\times10^{-17}$; the largest
equilibrium feasibility violation is $3.0\times10^{-9}$.
For the reversible case, actor bids $(1,3,5)$ and opponent bids $(1,2,3)$ give
\[
 M=\begin{pmatrix}1&-1&-1\\1&1&1\\0&1&1\end{pmatrix},\qquad v=1.
\]
Both opponent pure bids 1 and 2 are equilibrium mixtures, but the registered
actor contrast, bid 1 minus bid 5, is respectively $+1$ and $-2$.
Thus a specified equilibrium selector yields reproducible labels, while a
generic \emph{equilibrium opponent} does not uniquely order all candidate
actions. The two original model runs achieve only $30/47$ and $17/47$ on
the strict-direction subset, so target ambiguity does not explain all errors.
The development-only audit is reproduced by
\path{goofspiel_equilibrium_label_audit.py}; its JSON records every matrix
and extremizing mixture. Frozen labels, the $58/90$ and $38/90$ results, and
confirmation status remain unchanged.

\paragraph{Reported v0.4 development completion.}
The September 7 delivery report for commit \texttt{dd6735f} closes the next
development task. Initial adjacent-pair scoring reaches $48/90$ (53.3\%);
ranking all legal bids reaches $40/90$ (44.4\%). The single adjacent-pair
revision reaches $64/90$ (71.1\%), below the registered $75\%$ criterion.
Its new high/low slices are $29/42$ and $35/48$. The old v0.3 diagnostic
control gives $26/38$ on the high side, $13/52$ on the low side, and
$23/60$ on the large-gap subset; the low-side result is also below the
reported $30\%$ condition. All development runs report zero parser/API
failures and zero leakage. With the 1/1 revision budget exhausted, v0.4
confirmation remains sealed, the 30 v0.3 confirmation states remain unused,
and no formal cache or four-arm training is added. This is a development
stop, not an additional confirmation failure; v0.1 $7/29$ and v0.2 $18/29$
remain the frozen confirmation results.

The report also records Gate L0 success on $120/120$ reachable states and
nine passing unit tests. These v0.4 details are delivery-reported because
the corresponding raw increment was not included in the inspected package.
The primary record is \path{goofspiel_reward_v04_development_report.md},
with acceptance, candidate-selection, exposure, and split-audit metadata.
The v0.3 equilibrium audit above does not establish v0.4 label behavior.

\FloatBarrier
\section{Terminal Replay Certificates and Trajectory Multiplicity}
\label{app:terminal-replay}

\paragraph{Source and scope.}
The audit reads the 12 per-seed, per-arm episode files from the supplied
\texttt{cf23801} snapshot. Original bytes and input hashes are retained;
the later \texttt{f04b802} packaging completion is delivery-reported.
Independent grid and bitboard detectors agree on wins and winning-action
alternatives across 11,513 legal transitions and 720 terminal outcomes.
The checks include masks, player alternation, final actions, and fallback
counts. No training, model call, or policy rollout is added.

\paragraph{A terminal certificate.}
For a decision state $s$, let $W(s)$ be the legal actions whose deterministic
successor is an immediate win for the acting player. Define a wrapper that
preserves the recorded execution except that, on a fallback with
$W(s)\ne\varnothing$, it executes $\min W(s)$. For a recorded episode $i$,
let $C_i=1$ if this rule first changes an action along its original path.
Before that decision, the path is identical under fixed per-episode
exogenous randomness. At the changed decision, the replacement terminates
with a win. Therefore its binary outcome is exactly
\[
Y_i^{\mathrm{terminal}}=C_i+(1-C_i)Y_i,
\qquad
\overline{Y}^{\mathrm{terminal}}-\overline{Y}
=\frac{1}{n}\sum_i C_i(1-Y_i).
\]
The certificate requires prefix consistency and the specified wrapper.
Changing earlier actions or prompts defines a different counterfactual. The
post-hoc, rule-assisted intervention certifies local repairability; it does
not estimate decoder generalization or a training reward effect.

\begin{table}[ht]
\caption{Recorded outcomes and post-hoc terminal replay, each out of 60
games. The same fallback-only immediate-win rule is applied to both arms.
Replay entries are certified counterfactual totals, not additional empirical
policy evaluations.}
\label{tab:terminal-replay}
\centering\small
\begin{tabular}{lrrrr}
\toprule
Training seed & Policy recorded & Policy replay & Hybrid recorded & Hybrid replay\\
\midrule
20260800 & 28 & 28 & 18 & 28\\
20260801 & 19 & 19 & 19 & 29\\
20260802 & 28 & 28 & 28 & 28\\
\midrule
Total & 75 & 75 & 65 & 85\\
\bottomrule
\end{tabular}
\end{table}

The 122 Hybrid fallback actions include 40 missed immediate-win decisions
across 20 lost games; Policy's 20 fallback actions have no immediate-win
alternative. Each game has two opportunities, at zero-based steps 18 and 20, and is
counted once at step 18. The 20 first repairs repeat one board under two
checkpoints; all 40 opportunities involve two boards. These are the
denominators for diagnostic coverage: 20 certified episode prefixes, one
distinct first-repair board, and two affected checkpoints. The $20/20$
certificate coverage is a deterministic finite-archive result, with no
binomial confidence interval or estimate of repaired-policy win probability.
Valid Policy actions also miss immediate wins (84 decision visits across
the full archive). Thus the audit singles out a sufficient local fallback
repair without asserting that all tactical failures arise from parsing.

\paragraph{Multiplicity and the target of comparison.}
A joint trajectory class groups game slots only when their training seed,
opponent, seat, and all four complete executed player--action sequences
match. The 180 matched slots yield 72 classes: 60 singletons and 12 classes
with multiplicity ten. Uniformly weighting these 72 classes, instead of
the original game slots, gives Reward--Base $+4.17$ points, Hybrid--Policy
$-1.39$ points, and an interaction of $-5.56$ points. This changes the
estimand and is a sensitivity analysis, not a replacement leaderboard or
an estimate of effective sample size. Neither distinct trajectory classes
nor their within-class repeats are automatically independent samples.

For the Qwen discordance alone, the executed paired-path signature is
identical across all ten games. Counting $m$ copies of this one-sided
discordance produces nominal exact McNemar values of $1$, $0.5$, $0.0625$,
and $0.001953$ for $m=1,2,5,10$. This algebraic sensitivity demonstrates
why a small record-level $p$-value cannot establish behavioral diversity.
It supplies no corrected population-level significance claim.

\paragraph{Reproduction.}
The historical terminal-replay analysis used \texttt{analyze.py}, 12 raw
episode files, input hashes, a 720-row game audit, 40 decision certificates,
the ten discordant-pair records, and 72 trajectory classes. These complete
replay inputs and the historical analysis script are not included in the
current selected public evidence release. The released seed-summary verifier
reproduces the aggregate arithmetic; it does not independently reproduce
the terminal-replay certificates. The distinction is recorded in the public
release README.

\subsection{Exact Historical Seed Contrasts}
\begin{table}[tbp]
\caption{Exact descriptive historical contrasts, in percentage points. Initial seeds use 60 games per arm; added seeds use 180. Means weight training seeds equally within each batch. This table gives the numeric values underlying Figure~\ref{fig:historical-seed-contrasts}.}
\label{tab:historical-seed-contrasts}
\centering\small
\begin{tabular}{llrr}
\toprule
Batch & Training seed & Reward--Base & Hybrid--Policy\\
\midrule
Initial & 20260800 & $+1.6667$ & $-16.6667$\\
 & 20260801 & $+1.6667$ & $0$\\
 & 20260802 & $+1.6667$ & $0$\\
Initial mean & & $+1.6667$ & $-5.5556$\\
\midrule
Added & 20260803 & $+1.1111$ & $+15.5556$\\
 & 20260804 & $-1.1111$ & $0$\\
 & 20260805 & $+1.6667$ & $+16.6667$\\
 & 20260806 & $0$ & $-16.6667$\\
 & 20260807 & $-1.1111$ & $-17.2222$\\
 & 20260808 & $-1.1111$ & $0$\\
 & 20260809 & $0$ & $0$\\
Added mean & & $-0.0794$ & $-0.2381$\\
\bottomrule
\end{tabular}
\end{table}

\FloatBarrier
\section{Added Seven-Seed Batch and Loss-Consistency Audit}
\label{app:added-seeds}

\begin{table}[t]
\caption{Added seven-seed batch. Counts are wins out of 180 games per arm and
seed (no draws). Effects are percentage points; $I=(H-P)-(R-B)$. The final row
uses 1,260 games per arm. These are descriptive seed contrasts, not an
equivalence test.}
\label{tab:added-seed-results}
\centering\small
\setlength{\tabcolsep}{5pt}
\begin{tabular}{lrrrrrrr}
\toprule
Training seed & Base & Reward & Policy & Hybrid & $R-B$ & $H-P$ & $I$\\
\midrule
20260803 & 86 & 88 & 87 & 115 & $+1.11$ & $+15.56$ & $+14.44$\\
20260804 & 88 & 86 & 85 & 85 & $-1.11$ & $0.00$ & $+1.11$\\
20260805 & 84 & 87 & 84 & 114 & $+1.67$ & $+16.67$ & $+15.00$\\
20260806 & 85 & 85 & 115 & 85 & $0.00$ & $-16.67$ & $-16.67$\\
20260807 & 88 & 86 & 115 & 84 & $-1.11$ & $-17.22$ & $-16.11$\\
20260808 & 86 & 84 & 115 & 115 & $-1.11$ & $0.00$ & $+1.11$\\
20260809 & 86 & 86 & 85 & 85 & $0.00$ & $0.00$ & $0.00$\\
\midrule
Total / mean & 603 & 602 & 686 & 683 & $-0.08$ & $-0.24$ & $-0.16$\\
\bottomrule
\end{tabular}
\end{table}

\paragraph{Evidence inspected and budgets.}
The delivery is identified by commit \texttt{f6ed5e1148b0} and
\path{glm_v04_10seed_curated_evidence_index.tar.zst}. All 232 indexed file
hashes match. Checks cover training summaries, configurations, reload and
acceptance records, coverage reports, and complete episode files for seeds
20260803--20260809. Each of 28 training summaries records updates 1--100;
each evaluation file has 180 unique schedule slots with ten episodes per
evaluation-seed/opponent/seat combination. All opponent-level win totals
reproduce. Paired initialization hashes match within backbone; paired Qwen
adapter-file digests differ. Model reload was not independently executed.

Traditional summaries record 100 updates of 200 episodes, or 20,000 episodes
per run and 280,000 across 14 runs. Qwen uses 200 teacher episodes and 100
updates, with eight microsteps each. These are different data budgets, as
in the initial configurations. The new evaluation uses all three held-out
seeds per checkpoint, increasing games from 60 to 180. Results are reported
separately from the initial batch; the added schedule is not fixed-checkpoint
re-evaluation of the old weights.

The separate full archive reportedly contains 540 files in 1,719,146,441
compressed bytes. Its SHA-256 prefix is \texttt{b1b84b75b45e}; neither the
full digest nor the weight binaries were checked here. The fixed-checkpoint common-runtime re-evaluation has since been completed
and is reported in Appendix~\ref{app:runtime-identity}. The remaining
initial-batch gap concerns actual completed updates, example consumption,
and their checkpoint correspondence. The added batch has the numerical
update records described above, while full historical objective/source
correspondence remains unresolved. The two batches retain distinct training
and evaluation contracts and are analyzed separately.

\paragraph{Reward exposure.}
The curated coverage reports record Reward training hits of
$572{,}532/1{,}661{,}166=34.47\%$ and Hybrid training hits of
$5{,}846/13{,}705=42.66\%$. Evaluation hits are
$7{,}480/15{,}566=48.05\%$ and $7{,}000/19{,}555=35.80\%$,
respectively. All recorded train/evaluation API-call counters are zero.
These are lookup-based support rates, not nonzero-reward exposure or semantic
fidelity. Their denominators also differ from agent-only action counts,
so they are not substituted for the earlier effective-support audit.

\paragraph{Loss values versus the declared Qwen objective.}
The added Policy configuration has \texttt{advantage\_scale=1},
\texttt{reward\_weight\_min=0.25}, and \texttt{reward\_weight\_max=1.25}.
The summary and supplied source declare direct signed-advantage NLL, but
Table~\ref{tab:qwen-loss-consistency} shows that logged values do not follow
that declaration even in constant-outcome microbatches. At update 38, all
recorded outcomes in the batch are $-1$; its centered advantage is negative,
whereas its logged loss is positive. At the first update of each original
Policy run, the ratio is also approximately $1.25$, rather than the recorded
advantages $0.3751$, $0.3000$, and $0.3754$.

\begin{table}[ht]
\caption{Numerical Qwen loss audit: seed 20260803 Policy. Under the historical signed
$\mathrm{loss}=\operatorname{mean}(\widehat A\,\mathrm{NLL})$, a batch with
constant outcome and a common centered advantage would have loss/NLL equal
to that advantage. The logged ratio instead remains positive.}
\label{tab:qwen-loss-consistency}
\centering\small
\begin{tabular}{rrrrrr}
\toprule
Update & Outcome mean & Advantage mean & NLL mean & Loss & Loss/NLL\\
\midrule
1 & $+1$ & $+0.300268$ & $1.140075$ & $1.425093$ & $+1.25$\\
38 & $-1$ & $-1.699732$ & $1.388736\times10^{-5}$ & $3.471841\times10^{-6}$ & $+0.25$\\
\bottomrule
\end{tabular}
\end{table}

The independent recomputation covers all 100 logged updates for each of
14 Qwen runs. All 1,400 loss/NLL ratios are positive, with range
$0.25$--$1.2500000522$; the upper excess is floating-point error. This supports
the follow-up correction from signed-advantage language to positive bounded
effective weighting in the logs. In mixed batches the ratio is an NLL-weighted
aggregate, not an identified per-example coefficient. It does not determine
the full return-to-weight mapping or prove that logging equals backpropagation.

\paragraph{Follow-up source correspondence.}
The follow-up delivery reports GitHub commit \texttt{e16299f02d95}; the
archived training and evaluation results are unchanged.
The 2026-09-06 correction reports the historical runtime identities separately:
seed 20260803 training uses \texttt{ce93fb}, later added seeds use
\texttt{2fd323}, and the first added Hybrid evaluation also records
\texttt{f458d5d} in its environment field. The historical \texttt{2fd323}
source-file digest is explicitly unknown. The report describes a current
implementation using $w=\operatorname{clamp}(\operatorname{minmax}(R),0,1)$;
that formula alone cannot explain the historical ratio $1.25$.

The accompanying source snapshot still contains
\texttt{loss = nll * advantage / args.grad\_accum}. Its reward-objective
module digest begins \texttt{8624b26cc59e}, whereas the report gives
\texttt{8c6e4f8bf20a}; the loader prefixes are \texttt{8aa012c89687}
and \texttt{9a3358608abc}, respectively. We therefore distinguish the
verified numeric correction from the reported current-source change.
The exact supplied-byte hashes and two original metric rows accompany the
paper in \path{evidence/objective_correction_20260906/}.

\paragraph{Integrity of the follow-up attachment.}
The embedded curated archive contains 234 files, with 232 indexed payload
hashes that independently match. The outer files named
\path{MANIFEST_added_seed.json} and \path{SHA256SUMS_added_seed.txt}, however,
both contain the README text rather than a manifest/checksum list. Thus the
validated embedded index is retained as the integrity basis; a newly corrected
234-entry checksum release is not independently established by this attachment.
The reported 12 passing targeted tests are delivery-reported, not rerun here.
The README also retains a contradictory sentence about traditional episodes;
the verified budget remains 100 updates times 200 episodes, or 20,000 per run.

\paragraph{Runtime and follow-up scope.}
The README lists \texttt{2fd323}, while seed 20260803 training and three arm
evaluations record \texttt{ce93fb}. Its completion record reports resuming
Hybrid evaluation after a provider/trace-path invocation-guard fix without
retraining; that evaluation records source \texttt{2fd323} and environment
\texttt{f458d5d}. The remaining six seeds record \texttt{2fd323} in both phases.
The source-label exception remains part of the historical provenance record,
but fixed-checkpoint common-runtime re-evaluation across all ten historical
training seeds is now complete and reported in Appendix~\ref{app:runtime-identity}.
A separate step-100 sign-control report is recorded in
Appendix~\ref{app:direction100}; it does not come from this
teacher-adaptation delivery. The clean Connect Four on-policy implementation
is documented as a verified training-mechanism reference in
Appendix~\ref{app:training-references}; the Leduc on-policy extension is
reported separately in Appendix~\ref{app:leduc-policy}. The only unresolved
historical recovery required for budget matching is the original Policy/Hybrid
completed-update and sample-consumption record mapped to checkpoint identity.

\FloatBarrier
\section{Training-Mechanism References: SFT and On-Policy Connect Four}
\label{app:training-references}

\subsection{SFT-only reference}
A plain teacher-action cross-entropy reference uses the frozen source pools
without reward or advantage weighting. Each of three seeds consumes 80
examples from the beginning of its frozen source pool, performs 10 optimizer steps (80 microsteps), and is evaluated on 80
games. Total wins are 17, 31, and 18, giving $66/240=27.5\%$. On the shared
random\_v1, heuristic\_v1, and heuristic\_v2 subset the score is
$66/180=36.67\%$; the additional alpha-beta-d6 block is $0/60$. Per-opponent
wins are 49/60, 17/60, 0/60, and 0/60. The frozen source pools contain
1,125/1,120/1,124 examples, but only 80 per seed are actually consumed by
this SFT run.

This comparison is intentionally labeled a training-mechanism reference. It
does not establish a matched historical budget because the original
Policy/Hybrid completed updates and example-consumption order have not been
recovered.

\subsection{Normalized On-Policy Training and Formal-Checkpoint Evaluation}
The clean Connect Four experiment uses a legal-action categorical policy.
For permitted observation $o$ and legal set $\mathcal A(o)$, the model scores
each candidate completion, giving sequence log-scores $s_\theta(o,a)$, and uses
\begin{equation}
 \pi_\theta(a\mid o,\mathcal A(o))
 =\frac{\exp s_\theta(o,a)}{\sum_{b\in\mathcal A(o)}\exp s_\theta(o,b)}.
 \label{eq:c4-normalized-policy}
\end{equation}
The sampled action's normalized log-probability is also used in the update;
the training/evaluation scoring context limit is 512 tokens. Unlike the
historical free-text policy wrapper, this evaluator samples a legal candidate
directly. Parsing and fallback rates from that earlier wrapper are therefore
not measures of the new policy's execution reliability.

Policy and Hybrid each complete seeds 20260800--20260802, with 200 episodes
and 100 updates per run. The training records include parameter-change,
zero sampled/recomputed pre-update log-probability discrepancy, and checkpoint
reload/terminal-rollout checks. Hybrid uses the frozen v0.4 dense cache;
reported lookup rates are 28.93\%, 32.93\%, and 30.07\%, with zero judge API
calls and explicit misses. These query-based rates include their logged
acting-player scope and are not substituted for agent-only nonzero
optimization exposure. Neither this new training nor SFT reconstructs the
missing historical initial-batch sample/update record.

\paragraph{Formal checkpoint identity.}
The final 120-game evaluation directly loads the six original formal
training checkpoints identified by \path{FORMAL_ONPOLICY_MANIFEST.json}.
\path{CHECKPOINT_CORRESPONDENCE.json} maps each arm/seed to the formal
manifest hash and the evaluator's observed adapter hash; these match in all
six cases. The similarly named \path{LocusRL_final_results_20260819_v2}
checkpoint batch is excluded. The delivery records no conversion or re-save
of the formal adapters. The lightweight evidence permits correspondence
checks but does not contain the adapter binaries for an independent weight
rehash or GPU inference rerun.

\paragraph{Paired evaluation schedule.}
Each checkpoint plays ten games against each of \texttt{random\_v1} and
\texttt{heuristic\_v1}, five in each seat. Thus there are 20 games per
seed/arm, 60 per arm, and 120 overall. For training seed $s$, opponent index
$k\in\{0,1\}$ in that order, and episode ID $j\in\{0,\ldots,9\}$,
\begin{align}
 e(s,k,j)&=10^6s+700000+10000k+j,\\
 u(s,k,j)&=10^8s+7000000+10000k+j.
\end{align}
Here $e$ is the environment seed and $u$ the explicitly set PyTorch sampling
seed; seat is $j\bmod2$. Policy and Hybrid use the same environment/opponent
schedule and paired action-sampling seeds. The delivery separates this
schedule from training. Shared seeds fix the evaluation contract; they do
not force different policies to visit identical states.

\begin{table}[tbp]
\caption{Formal on-policy evaluation by opponent and seat. Every row is a
slice of the same 120-game evaluation, not another experiment. There are no draws.}
\label{tab:onpolicy-formal-slices}
\centering\small
\begin{tabular}{lrr}
\toprule
Slice & Policy wins/games & Hybrid wins/games\\
\midrule
random\_v1 & 24/30 & 19/30\\
heuristic\_v1 & 15/30 & 14/30\\
\midrule
Seat 0 & 27/30 & 24/30\\
Seat 1 & 12/30 & 9/30\\
\bottomrule
\end{tabular}
\end{table}

The seed-wise results are in Table~\ref{tab:onpolicy-formal-results}. The
60 paired episode slots contain 30 joint wins, 18 joint losses, nine
Policy-only wins, and three Hybrid-only wins. For training-seed inference,
Hybrid-minus-Policy contrasts are $-10$, $-20$, and $0$ pp. Omitting the tie,
the exact two-sided sign test gives $p=0.5$ under independent equiprobable
nonzero signs. The three training blocks, rather than 60 episode pairs,
define the replication scale. Neither general superiority nor equivalence
is established by this small experiment.

\paragraph{Artifact verification.}
The formal delivery is retained under
\begin{quote}\small
\path{evidence/connect_four_onpolicy_eval_formal_20260921/}.
\end{quote}
It contains \path{summary.json}, \path{episodes.jsonl},
\path{VALIDATION.json}, \path{CHECKPOINT_CORRESPONDENCE.json}, and
\path{README.md}, with five matching entries in
\path{FINAL_SHA256SUMS.txt}. Independent recomputation recovers all arm,
seed, opponent, and seat totals. Replaying all 120 games and 1,106 actions
checks pre-action boards, legal sets, alternating players, executed moves,
terminal outcomes, and returns without discrepancies. Environment and Torch
seed formulas and all 60 pairings also reproduce. This verifies the recorded
trajectories and schedule, separately from the delivery's real-model training
and reload checks.

The model archive is identified by
\path{LocusRL_formal_onpolicy_checkpoints_20260921.tar.gz}, with SHA-256
\begin{quote}\small\ttfamily
7ab893bc7ae43d2cdb429ae371dffb5e9c\\
74f1426b71d73d5cef0642d9ffa6fc.
\end{quote}
The delivery records instance-local storage rather than an independently
verified persistent download location. Artifact checksums establish identity;
they do not by themselves establish long-term weight availability.

\section{Verification Status and Historical Recovery}
\label{app:verification-targets}

\paragraph{Completed evidence.}
The paper includes the ten-seed fixed-checkpoint common-runtime evaluation,
1,400 logged-update numerical audit, three-seed step-100 reward-direction
control, SFT mechanism reference, normalized Connect Four on-policy training
and formal-checkpoint 120-game evaluation, terminal-replay certificates,
Leduc fresh-RNG comparison, and reward-reference/exposure checks. These
experiments retain separate learners, budgets, evaluation schedules, and
replication units.

\paragraph{Historical provenance.}
For the initial Policy/Hybrid batch, original training-instance records are
needed to establish completed optimizer steps, actual examples consumed,
sampling/consumption order, and their mapping to each checkpoint. The
configuration \texttt{TRAIN\_MAX\_STEPS=10} establishes a requested budget,
not completed training. Unrecoverable fields are reported as unavailable;
later SFT or on-policy runs cannot supply historical execution evidence.
The added batch already contains 100 logged updates for each of 14 Qwen
runs. Its complete return-to-weight and backpropagation implementation
remains a separate source-correspondence question, as detailed in
Appendix~\ref{app:added-seeds}.

\paragraph{Weight availability and extension scope.}
The formal on-policy delivery identifies an instance-local checkpoint
archive and per-adapter hashes. Independent persistent availability is not
established by the lightweight bundle. This storage limitation is separate
from the completed training, evaluation, and trajectory checks.
Prospective evaluation of a deployed immediate-win fallback wrapper and
measurement of the protocol's diagnostic performance on new systems would
extend the present findings; they are not counted as completed experiments.

\begin{table}[tbp]
\caption{How an additional diagnostic check changes interpretation in the inspected cases. These are case findings, not detector-accuracy trials.}
\label{tab:diagnostic-yield}
\centering\footnotesize
\begin{tabularx}{\linewidth}{p{0.16\linewidth}YY}
\toprule
Object & Initial summary & Interpretation after the check\\
\midrule
Reward semantics & 9,954 structurally valid records & Reference revalidation separates avoidable error from forced loss; positive scores occur in both classes\\
Replication & Near-zero added-batch mean & Opposing seed-wise contrasts are hidden by the mean\\
Evaluation & Different historical runtime labels & Fixed checkpoints have directly measured outcomes under a common contract\\
Optimization & Declared signed teacher-action update & Positive logged loss/NLL ratios contradict that description; a separate normalized on-policy experiment is trained and evaluated\\
Reward influence & Dense shaping is present & Sign reversal strongly changes learning; original-sign shaping does not exceed terminal-only training\\
Execution & Final actions are legal & A repeated fallback misses an immediate win; terminal replay verifies a sufficient correction\\
Validation & Current splits are disjoint & Historical exposure prevents independent confirmation; equilibrium selection can change action labels\\
\bottomrule
\end{tabularx}
\end{table}

\section{Diagnostic Workflow and Distribution Companion Views}
\label{app:visual-companions}

\begin{figure}[ht]
\centering
\includegraphics[width=0.96\linewidth]{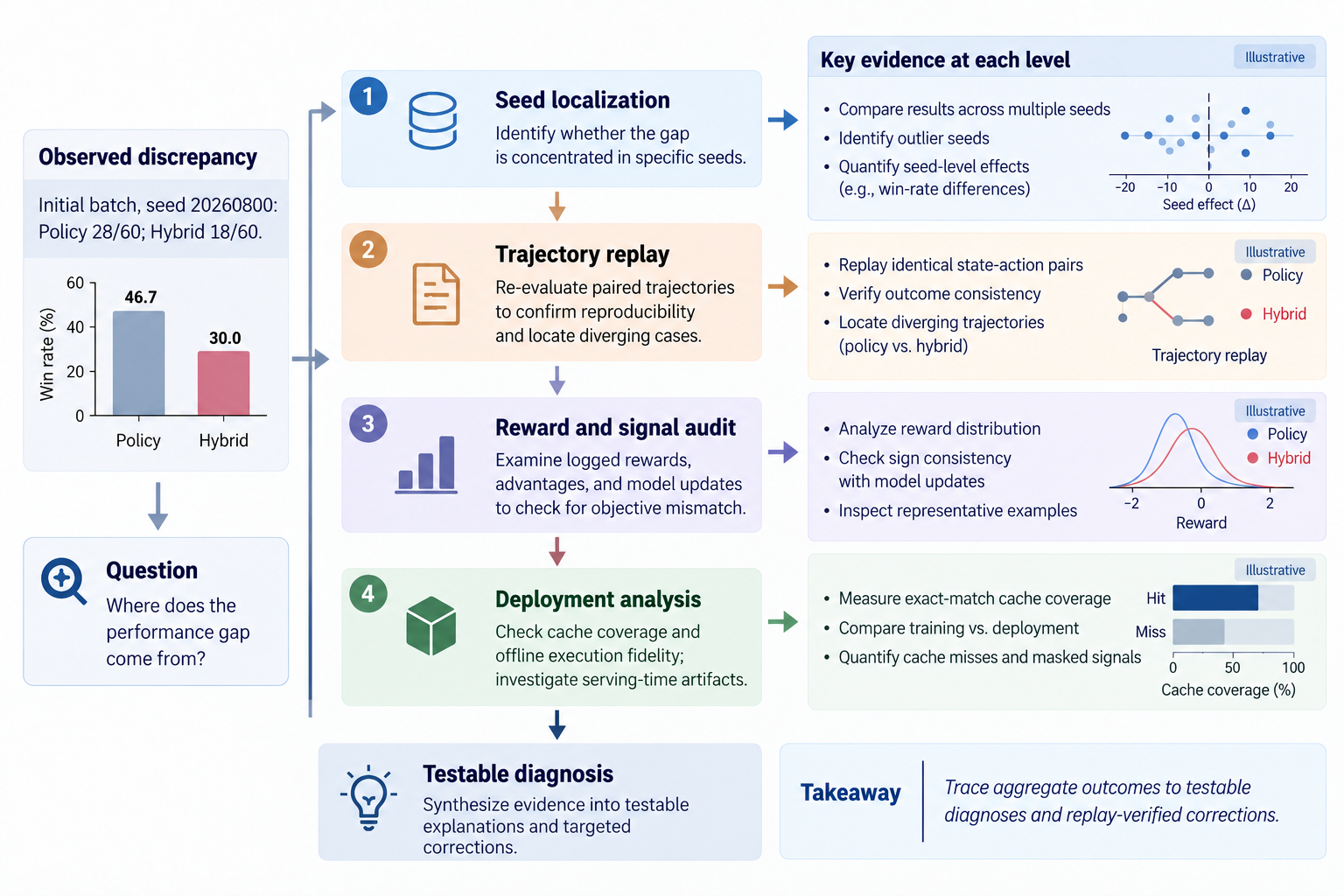}
\caption{Comparison-to-diagnosis workflow: seed localization, trajectory replay, reward-signal audit, and deployment checks. The leftmost $46.7\%$ versus $30.0\%$ comparison is Policy versus Hybrid for initial seed 20260800 ($28/60$ versus $18/60$), not the whole batch. Small plots illustrate the checks and add no measured results. Table~\ref{tab:main-protocol} gives the operational decision logic.}
\label{fig:diagnostic-workflow}
\end{figure}

The detailed deployment diagram in Figure~\ref{fig:system-overview} explains how the frozen reward reaches each policy system. Figure~\ref{fig:diagnostic-workflow} provides the complementary comparison-to-diagnosis view. These are two views of the same protocol, with the optimization-identity check separating the intended design from its realized implementation.

\begin{figure}[ht]
\centering
\includegraphics[width=0.98\linewidth]{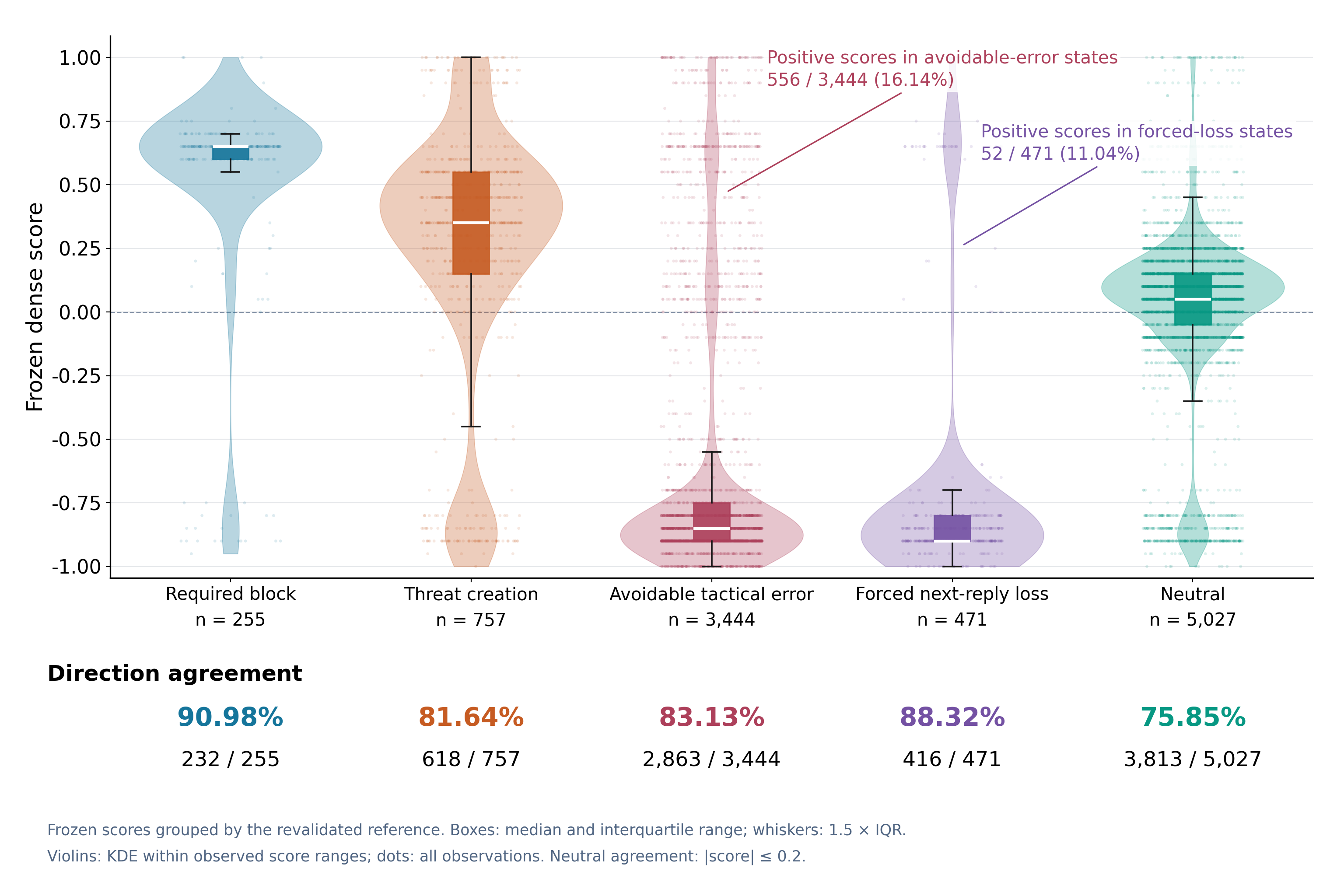}
\caption{Distribution companion under the five-category revised reference. Violin, box, and scatter layers visualize frozen scores; exact category counts and agreement rates are given in Table~\ref{tab:glm-cache-audit}. Positive scores occur in $556/3{,}444$ avoidable-error records and $52/471$ forced-loss records. The latter are reference-direction disagreements in unavoidable-loss states, not evidence that a safe action was available. Plot smoothing and scatter placement are visualization choices, not additional observations.}
\label{fig:reward-distribution-detailed}
\end{figure}

Figures~\ref{fig:reward-distribution-detailed} and~\ref{fig:reward-category-distribution}
provide distribution and score-sign views of the same 9,954 frozen scores after
reference revalidation. Table~\ref{tab:glm-cache-audit} contains the canonical
counts; no frozen score is changed by the relabeling.

\section{Reference Transitions and Seed Localization}
\label{app:original-figure-layouts}
These companion layouts summarize reference transitions and the initial
three-seed contrasts. They reuse existing records; the canonical reference
counts remain in Table~\ref{tab:glm-cache-audit}.

\begin{figure}[ht]
\centering
\includegraphics[width=0.94\linewidth]{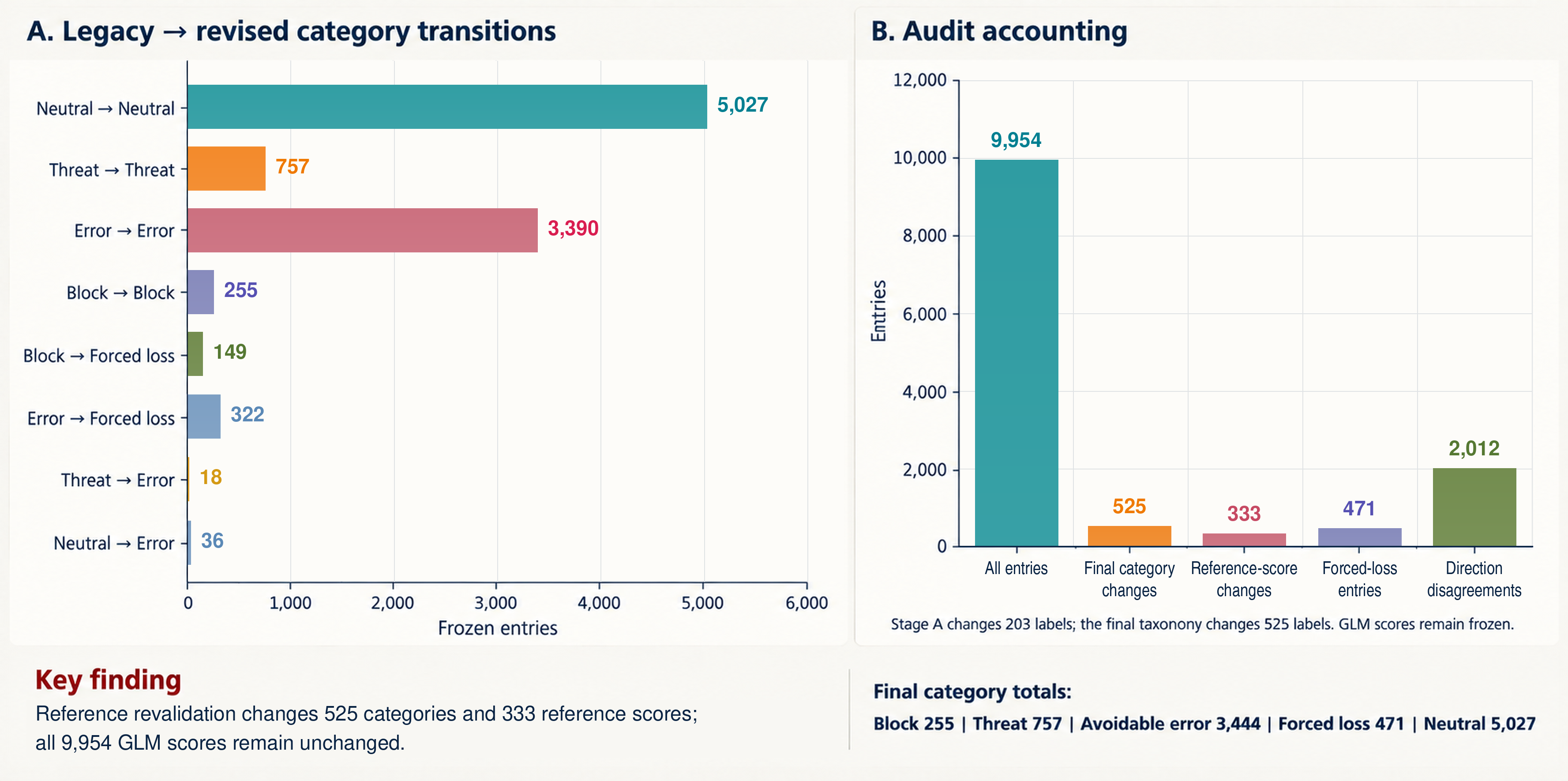}
\caption{Reference-revalidation accounting. Panel A shows legacy-to-revised category transitions; Panel B reports 525 final category changes, 333 reference-score changes, 471 forced-loss entries, and 2,012 direction disagreements. Stage A changes 203 labels before the final two-stage taxonomy reaches 525 changes. Final category totals are $255$, $757$, $3{,}444$, $471$, and $5{,}027$; $3{,}390$ is a transition count, not the final avoidable-error total. Frozen GLM scores are unchanged; Appendix~\ref{app:reference-revalidation} gives the accounting.}
\label{fig:original-distribution-layout}
\end{figure}

\begin{figure}[ht]
\centering
\includegraphics[width=0.98\linewidth]{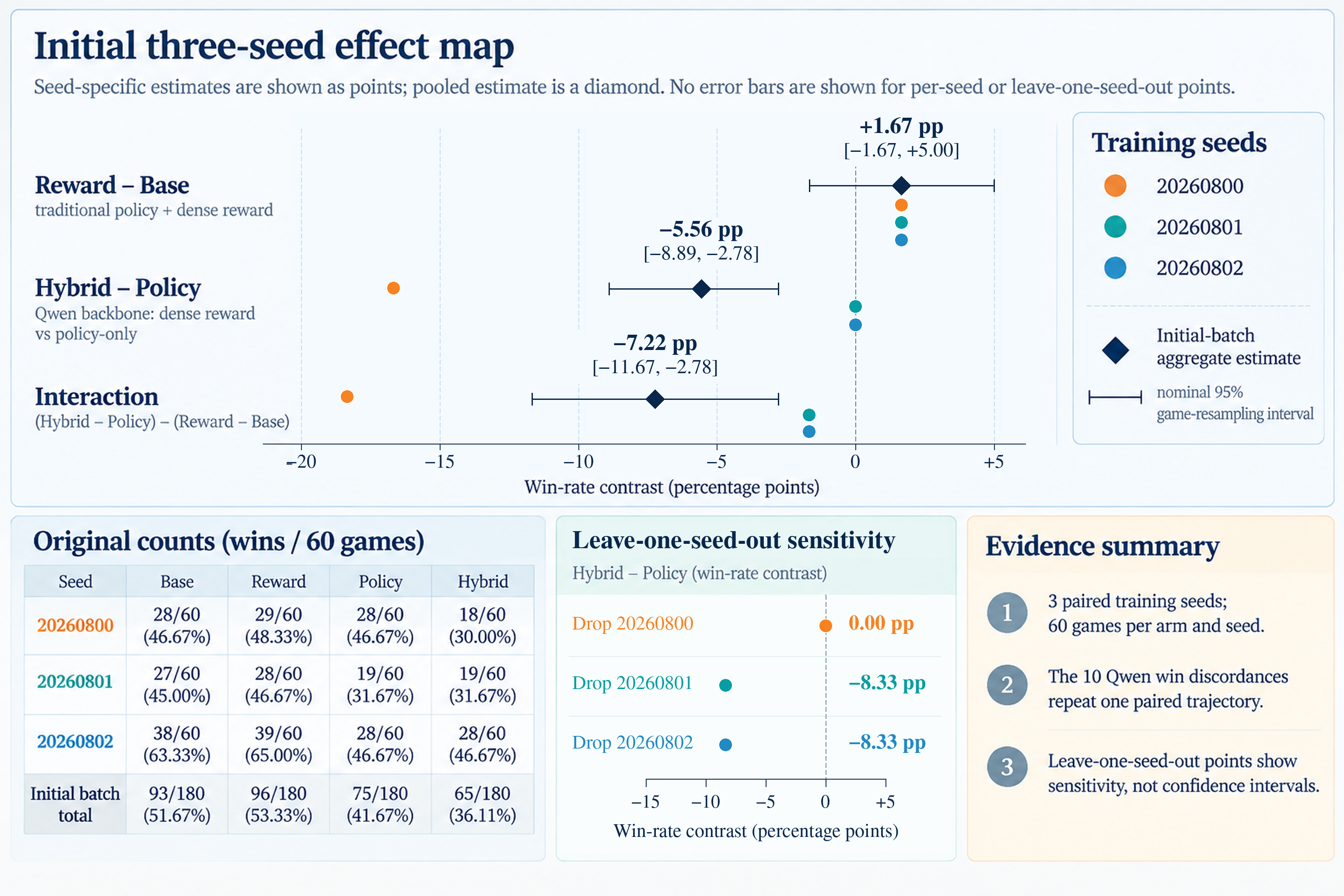}
\caption{Companion effect map for the initial three-seed batch. Labelled contrasts and win counts describe the initial results. Diamonds are initial-batch aggregate estimates with nominal 95\% game-resampling intervals; per-seed and leave-one-seed-out points are shown without local error bars. Leave-one-seed-out Hybrid--Policy estimates are $0$, $-8.33$, and $-8.33$ percentage points. Evaluation-label differences alone do not establish a runtime effect, and ten Qwen discordances repeat one paired trajectory. This view adds no independent experiment.}
\label{fig:original-effect-layout}
\end{figure}

\section{Reference Revalidation from Human Trial Counterexamples}
\label{app:reference-revalidation}

\subsection{Discovery Record and the Defective Predicate}

A ten-case human trial review of frozen Connect Four reward records identified
two high-confidence directional disagreements with the original reference
(\texttt{JH-TRIAL-03} and \texttt{04}) and one additional taxonomy correction
(\texttt{07}). The corresponding workbook and correction report preserve the
annotator's distinctions. The trial materials expose reference and GLM fields,
so these ten cases serve as development witnesses rather than a blind human
accuracy estimate. The subsequent 60-case submission is analyzed separately
in Appendix~\ref{app:formal60}. All ten reviewed state hashes, candidate actions,
acting players, and frozen scores match the cache audit.

The original \texttt{c4\_oracle.py} computes the opponent's winning replies
after the candidate but does not use that set to validate defense. Instead,
it sets \texttt{blocked} when the candidate occupies any previously winning
opponent column and defines an immediate loss as a pre-existing threat that
was not blocked. This can reward an incomplete defense and can also miss a
new winning reply enabled by the candidate's change to a column's height.
The frozen GLM rubric already requires a defensive move to remove the
opponent's immediate win and forbids rewarding a move that leaves an
immediate winning reply. The revision implements this check on the complete
successor state.

\begin{table}[ht]
\caption{Three distinct trial states verified by enumerating every legal
actor action. Columns are zero-indexed. The actor has no immediate win and
no safe action in any row. Cases 03 and 04 correct the reference's reward
direction; case 07 changes its category while retaining a negative direction.}
\label{tab:human-oracle-witnesses}
\centering\small
\begin{tabular}{lcrrrr}
\toprule
Trial & Player / action & Opponent wins before & Wins after & Old reference & GLM\\
\midrule
03 & $0/2$ & $\{2,6\}$ & $\{2,6\}$ & Block $+0.3$ & $-0.9$\\
04 & $1/2$ & $\{0,2\}$ & $\{0\}$ & Block $+0.3$ & $-0.9$\\
07 & $0/3$ & $\{2,4\}$ & $\{2,4\}$ & Error $-0.9$ & $+1.0$\\
\bottomrule
\end{tabular}
\end{table}

\paragraph{Complete next-reply witnesses.}
Writing $a\mapsto R$ for a legal actor move and its remaining opponent
winning columns, case 03 has
$\{0,1,2,5\}\mapsto\{2,6\}$ and $6\mapsto\{2\}$;
case 04 has $0\mapsto\{2\}$, $2\mapsto\{0\}$, and
$\{3,4,6\}\mapsto\{0,2\}$;
case 07 has $2\mapsto\{4\}$, $4\mapsto\{2\}$, and
$\{0,1,3,5,6\}\mapsto\{2,4\}$.
Every listed reply set is nonempty. In case 03, playing column 2 enables a
new landing cell in the same column, so its post-action winning set is still
$\{2,6\}$. Subtracting the chosen column from the pre-action set would also
be incorrect. The JSON audit includes the three boards and these complete
action--reply maps. Cases 03 and 04 retain the GLM score $-0.9$ as consistent
under the corrected reference; case 07's $+1.0$ remains inconsistent.

\subsection{Successor-State Rule and Forced-Loss Separation}

Let $A(s)$ be the legal columns, $W_p(s)$ the actor's immediately winning
actions, and $R(s,a)$ the opponent's immediately winning actions after the
actor plays a nonwinning $a$. Stage A replaces the original loss predicate
with $R(s,a)\ne\varnothing$. A required-block label is assigned only when a
pre-action opponent threat exists, the candidate is not an actor win or
missed win, and $R(s,a)=\varnothing$. The remaining precedence and magnitudes
are retained: immediate win $+1.0$, missed win with losing reply $-1.0$,
missed win $-0.8$, losing reply $-0.9$, required block $+0.3$,
threat/fork $+0.6/+0.8$, and neutral $0$.

Stage B separately labels the finite, next-reply condition
\begin{equation}
 \operatorname{ForcedLoss}(s) =
 \mathbf{1}\!\left[ W_p(s)=\varnothing\ \land\
      \forall a\in A(s),\ R(s,a)\ne\varnothing \right].
 \label{eq:forced-loss}
\end{equation}
The human proposal's two-or-more opponent winning columns provide a
sufficient witness when the actor has no immediate win. Equation~\ref{eq:forced-loss}
also covers cases revealed only by complete successor enumeration. It
identifies unavoidable loss on the opponent's next move, not arbitrary
long-horizon minimax loss. The negative reference score $-0.9$ in this class
describes that forced outcome, not causal blame for the selected action;
all legal actions lose under the stated one-reply condition. Other
\texttt{obvious\_error} entries retain an available immediate win or a move
that avoids an immediate losing reply, and are reported as avoidable
tactical errors.

Score-direction rules remain positive for win/block/threat, negative for
error/forced loss, and $|f|\leq0.2$ for neutral. The distinct historical
threshold test remains $f\geq0.2$ for a win, $f\geq-0.4$ for block/threat,
$f\leq-0.2$ for error/forced loss, and $|f|\leq0.6$ for neutral. Stage B
therefore changes the taxonomy without changing Stage A's agreement count.

\subsection{Exhaustive Reanalysis and Exact Accounting}

The reanalysis joins 9,954 unique state--action--player keys at 9,181 distinct
states. Before applying any revision, the verifier reproduces every original
category, reference score, direction flag, and threshold flag. It checks
gravity, turn counts, legal columns, nonterminal source boards, and all
canonical state hashes. Independent grid scanning and bitboard calculations
agree across 474,880 successor and reply computations.

\begin{table}[ht]
\caption{Exact reanalysis of fixed scores. Stage A corrects successor checks;
Stage B additionally separates next-reply forced loss. Agreement counts use
the same score-direction convention throughout.}
\label{tab:reference-reanalysis}
\centering\small
\begin{tabular}{lrrrr}
\toprule
Reference & Agreement & Disagreement & Threshold pass & Forced-loss class\\
\midrule
Original & $7{,}867$ & $2{,}087$ & $8{,}498$ & --\\
Stage A & $7{,}942$ & $2{,}012$ & $8{,}567$ & --\\
Stage B & $7{,}942$ & $2{,}012$ & $8{,}567$ & $471$\\
\bottomrule
\end{tabular}
\end{table}

Stage A changes 203 labels under the pre-separation taxonomy: 149 originally required blocks,
18 threats, and 36 neutral moves become errors. Stage B identifies 471
forced-loss entries at 469 distinct states: the 149 former blocks and 322
original errors. Of those entries, 413 already have at least two opponent
winning columns; the other 58 satisfy the full successor condition.
The final taxonomy changes 525 entries in total. Reference score magnitudes
change on 333 entries, including 130 already labeled as errors; the union of
category or magnitude changes contains 655 entries. None of these operations
changes a stored GLM score.

There are 135 old disagreements that become agreements and 60 old agreements
that become disagreements, leaving a net reduction of 75 disagreements.
The remaining 1,952 disagreements persist. Revised agreement is
$7{,}942/9{,}954=79.79\%$. Positive rewards occur on 556/3,444 avoidable
errors and 52/471 forced-loss entries; pooling these two classes would hide
the distinction between a remediable choice and an already forced outcome.
These counts are a census of the frozen cache, not independent human trials
or an estimate of performance on unseen states. The three trial witnesses
are distinct boards; training-seed replication remains as specified in
Appendix~\ref{app:statistical-diagnostics}.

\subsection{Provenance and Reproduction}

The standalone \path{analysis/recheck_c4_reference.py} consumes the frozen
pool and original score-audit trace, checks their full SHA-256 values, and
writes \path{analysis/c4_reference_audit_20260908.json}. It uses the Python
standard library and performs no model calls. The output preserves all old
and revised count tables, every changed entry, and the ten trial boards with
their complete legal-action replies. With the two frozen inputs available:
\begin{verbatim}
python analysis/recheck_c4_reference.py \
  --pool records.jsonl \
  --audit-trace full_cache_audit_trace.jsonl \
  --output analysis/c4_reference_audit_20260908.json
\end{verbatim}
The pool is the repository's
\path{c4_llm_reward_v03_rollout_pool_v3/records.jsonl} and has SHA-256:
\begin{center}\small\texttt{d7c704625fb158bd96fe0b6aa621509c9f}\linebreak
\texttt{a2d559bfc656825201c8721f9cba6f}.\end{center}
The original \path{full_cache_audit_trace.jsonl} has SHA-256:
\begin{center}\small\texttt{55e67b659bc2458d834cf023bf91d67b7}\linebreak
\texttt{aae769feac22654b4368c5635d5379d}.\end{center}
Cases 03, 04, and 07 use source keys \path{c4pool-v3-6609},
\path{c4pool-v3-5823}, and \path{c4pool-v3-2183}, respectively.
The output includes their canonical state hashes and boards.

This is a post-hoc correction of the full-cache reference. It does not rerun
the historical calibration screen, change the deployed reward artifact, or
alter training and evaluation outcomes. Lookup counts are label-independent;
the original evaluation semantic-support join remains explicitly historical
in Table~\ref{tab:evaluation-semantic-support}. The final supplied reward
figures use the revised category grouping; the numerical evidence is in
Tables~\ref{tab:glm-cache-audit} and~\ref{tab:reference-reanalysis}.

\section{Sixty-Case Annotation and Protocol Comparison}
\label{app:formal60}

\subsection{Materials, Provenance, and Replication Units}

The submitted first-stage set contains \texttt{JH-FORMAL-001} through
\texttt{060}: 60 distinct state--action--player records at 60 distinct boards.
Horizontal-reflection deduplication still leaves 60 boards, and none of their
board arrays occurs in the ten-case trial set. These are purposively selected
records from the existing frozen cache, rather than a new random sample or
new training runs. They broaden case coverage without adding training-seed
replicates or terminal-repair trials. The ten forced-loss witnesses below are
already included in the full-cache census in
Appendix~\ref{app:reference-revalidation}; their counts are not added to it.

The first-stage package supplies the board, player, candidate action, and
legal columns, withholding Oracle and GLM fields. The project lead reports
that the annotator first completed manual labeling and then used AI for
confirmation. The received illustrated workbook and expanded workbook contain
the same 60 labels, reward directions, and confidence values, with some
explanation wording differences. They are two presentations of one annotation
submission, not two annotators or two replications. A separate pre-AI snapshot
was not supplied, so the submission does not establish which answers, if any,
changed during AI confirmation. We preserve the received answers and keep
the subsequent computational audit distinct from human adjudication.

All cases contain a label, direction, confidence, and explanation. Submitted
directions are positive for 23 cases, negative for 27, and neutral for ten;
50 cases have high confidence and ten have medium confidence. The submitted
procedure assigns medium confidence to the neutral category, so these
confidence counts are not a calibrated uncertainty measure.

\subsection{A Precedence Conflict in the Issued Instructions}

The issued annotation order is immediate win, forced loss, required block,
obvious error, threat, then neutral. Its block predicate requires one
opponent winning column and a candidate that removes the threat completely.
The paper's reference instead checks for a missed actor win before assigning
a block label. These predicates overlap when a candidate successfully
defends but another action wins the game immediately.

Table~\ref{tab:formal60-precedence} lists all five affected cases. Each
submitted \texttt{must\_block} label follows the issued annotation protocol;
the paper's rule assigns \texttt{obvious\_error} with score $-0.8$ for missing
an immediate win. This precedence was already present in the legacy Oracle
and remains in the revised reference. The discrepancy is between the
annotation instructions and the paper, not a newly changed Oracle rule.
The first-stage answers remain intact; the harmonized second-stage protocol
checks a candidate win, then a missed actor win, before defense. The completed
review adopts this order for all five cases and records the AI-assisted
changes to \texttt{obvious\_error} / Negative with case-specific reasons.

\begin{table}[!htbp]
\caption{Five protocol conflicts verified directly from the boards.
All candidates completely block the opponent's immediate threat, but fail
to take an available immediate win. Columns are zero-indexed. Every submitted
label is \texttt{must\_block}; both legacy and revised Oracle labels are
\texttt{obvious\_error}.}
\label{tab:formal60-precedence}
\centering\small
\begin{tabular}{llrrrr}
\toprule
Formal case & Pool key suffix & Player & Candidate & Own winning columns & GLM\\
\midrule
006 & 9451 & 1 & 1 & $\{4\}$ & $+0.65$\\
015 & 1539 & 0 & 5 & $\{2\}$ & $+0.65$\\
030 & 8200 & 1 & 3 & $\{2\}$ & $+0.70$\\
032 & 3353 & 1 & 2 & $\{4\}$ & $+0.60$\\
060 & 2502 & 1 & 1 & $\{3,4\}$ & $+0.45$\\
\bottomrule
\end{tabular}
\end{table}

The pool key prefix is \texttt{c4pool-v3-}. Complete enumeration reproduces
all 60 submitted categories under the issued annotation precedence, whereas
55/60 match the paper's revised categories. These are comparisons between
explicit protocols, not estimates of human accuracy or inter-annotator
agreement. No first-stage answer is retroactively overwritten to improve
either count.

\subsection{Reference Revisions and Frozen-Score Comparison}

Ten other cases, 010, 021, 026, 028, 036, 038, 043, 054, 055, and 058,
receive \texttt{must\_block} from the legacy Oracle but
\texttt{forced\_loss} from both the submitted annotations and the complete
successor rule. They have no actor immediate win, and every legal action
leaves an immediate winning reply. Their revised score is $-0.9$; it
describes the forced outcome rather than attributing failure to the selected
action. Table~\ref{tab:formal60-categories} separates this reference correction
from the five precedence conflicts.

\begin{table}[!htbp]
\caption{Category accounting for the same 60 selected records.
Submitted labels use the issued annotation protocol; reference columns are
computed. Final labels come from completed AI-assisted review after the
comparison was disclosed.}
\label{tab:formal60-categories}
\centering\small
\begin{tabular}{lrrrr}
\toprule
Category & Submitted & Legacy Oracle & Revised reference & Final review\\
\midrule
Immediate win & 0 & 0 & 0 & 0\\
Required block & 15 & 20 & 10 & 10\\
Obvious error & 17 & 22 & 22 & 22\\
Forced loss & 10 & 0 & 10 & 10\\
Create threat & 8 & 8 & 8 & 8\\
Neutral & 10 & 10 & 10 & 10\\
\midrule
Total & 60 & 60 & 60 & 60\\
\bottomrule
\end{tabular}
\end{table}

Using the paper's direction rule (strictly positive for win/block/threat,
strictly negative for error/forced loss, and $|f|\leq0.2$ for neutral),
the frozen GLM scores agree with the legacy Oracle in 31/60 cases and with
the revised reference in 37/60. Eight disagreements become agreements and
two agreements become disagreements; 29 agreements and 21 disagreements
persist. The 23 revised-reference disagreements include all five precedence
conflicts, two forced-loss cases (010 and 038, both scored $+0.65$), and
16 other cases. These are descriptive counts for the selected set and cannot
be extrapolated to whole-cache or unseen-state accuracy.

The review queue places 31 distinct cases first: five protocol conflicts,
ten reference revisions, and 16 remaining GLM disagreements. Thus the
priority groups do not double-count the seven GLM disagreements already in
the first two groups. The other 29 cases remain available for routine review.
The returned second-stage workbook completes all 60 decision, final-category,
direction, and reason records: every decision is \texttt{Accept analysis}
and every status is \texttt{Recorded}. Final labels match the revised
reference in all 60 cases, with 18 positive, 32 negative, and ten neutral
directions. Only cases 006, 015, 030, 032, and 060 change from their submitted
labels and directions; each reason explicitly records AI assistance and the
missed immediate win. The ten forced-loss cases retain their first-stage
negative judgments. This completes adjudication under the harmonized
protocol. Because the reviewer saw the computed evidence and used AI, the
60/60 acceptance is a review outcome, not a blinded independent accuracy
estimate or inter-annotator agreement. Frozen GLM agreement remains 37/60;
review acceptance does not improve the scores. Original GLM explanations
were not archived, so the comparison evaluates score direction rather than
the model's unobserved reasoning.

\subsection{Audit Trail and Archived Inputs}

Each case is joined to the frozen pool by its board, acting player, and
candidate action, then to the score trace through its pool key. Player,
candidate, and legal-column fields match the submitted workbook. Independent
grid scanning and bitboard calculations agree in 2,992 candidate and reply
checks across these 60 cases. The preserved expanded submission, the full
comparison, the harmonized protocol, and the human-review queue are sheets
in \path{LocusRL_60_Case_Stage2_Comparison_20260908.xlsx}. The comparison
retains pool keys, trace entry identifiers, canonical state hashes, frozen
scores, winning columns, and safe or winning alternatives. Its human-review
sheet embeds the original board images.

The first-stage \path{cases_blind.jsonl} SHA-256 matches the submitted
production report:
\begin{center}\small
\texttt{03449602c9af0132bb785a3f5ecf8a563}\linebreak
\texttt{e2c63695db6c26c88a3fd6623d3431c}.
\end{center}
The primary illustrated submission has SHA-256:
\begin{center}\small
\texttt{90ebfc10fde46b0a66f207b13f1289129}\linebreak
\texttt{dbeebb9194f11c16c108af58adaafe4}.
\end{center}
The archived second-stage workbook, before human adjudication, has SHA-256:
\begin{center}\small
\texttt{e014f5abda7525e93f7465182cf23c62}\linebreak
\texttt{0724bb7814283ee7db41e9fab5e9240b}.
\end{center}
The workbook's protocol sheet also records both submitted workbook hashes,
the production-report hash, and the pool, trace, and reference-implementation
hashes. The full-cache inputs and executable reference definitions are those
in Appendix~\ref{app:reference-revalidation}.

The returned completed-review workbook has SHA-256:
\begin{center}\small
\texttt{f00f14efbd9be9d0db4555e333c72f6e}\linebreak
\texttt{eeba65692441419550a1b29f91659f36}.
\end{center}
Comparison against the archived pre-review workbook verifies unchanged
cell values and formulas in \texttt{60 Cases} and \texttt{Comparison}.
\texttt{Protocol} retains its content, with its cached completed-review count
updated from zero to 60. Changes in \texttt{Review} are confined to the
60 rows' decision fields D--G and their cached status H. All 428 formulas
are unchanged and their stored results match independent recomputation.
All 60 embedded PNG files are byte-identical to the archived originals,
with the same image-to-row placements. The preserved introductory
``pending'' text in \texttt{Comparison} describes the pre-review snapshot;
completed decisions are recorded in \texttt{Review}. These checks preserve
the distinction between original answers, computed evidence, and subsequent
AI-assisted adjudication.

\paragraph{Thirty-case supplemental review.}
A later 30-case supplement used a different workflow: AI produced a first-pass
analysis and a human reviewer then checked every case individually. All 30
were accepted after that review. This is reported as human confirmation of an
AI-assisted first pass, not as an independent blind accuracy estimate or
inter-annotator agreement measure, and it is kept separate from the preceding
60-case human-first workflow.

\section{Verified Step-100 Historical-Reward Direction Control}
\label{app:direction100}

\subsection{Design and Completed Artifact Audit}
A separate experiment compares outcome-only,
original-sign historical-oracle, and sign-flipped-oracle learning for training seeds
20260800--20260802, archived under reported commit \texttt{336d928} from
base \texttt{dd6735f}. Each condition uses an MLP learner, 100 updates and
200 newly sampled episodes per update (20,000 training games per seed).
The objective combines terminal outcome with the agent's subsequent dense
rewards, with discount and dense scale both one. Initialization tensor
storage and the complete step-0 evaluation traces match within each seed.
The providers satisfy $f_-(s,a,p)=-f_+(s,a,p)$ on matched inputs, with the
same terminal-reward handling.

The artifact audit verifies all 94 listed hashes, 1,482,846 optimization
samples, 54 learning-curve cells, and the 54 checkpoint hashes. It replays
all 3,240 evaluation games and 42,191 actions, also checking outcomes with
an independent board scanner. These checks establish record and source
consistency; they do not rerun training or checkpoint inference.

\begin{table}[!htbp]
\caption{Verified final-step historical-reward direction control, step 100. Each
seed--condition cell has 60 evaluation games. This batch is separate from
both the formal GLM four-arm runs and the historical five-update control.}
\label{tab:direction100}
\centering\small
\begin{tabular}{lrrr}
\toprule
Training seed & Outcome only & Positive oracle & Negative oracle\\
\midrule
20260800 & $38/60$ & $30/60$ & $1/60$\\
20260801 & $27/60$ & $27/60$ & $1/60$\\
20260802 & $38/60$ & $27/60$ & $3/60$\\
\midrule
Total wins & $103/180$ & $84/180$ & $5/180$\\
Win rate & $57.22\%$ & $46.67\%$ & $2.78\%$\\
\bottomrule
\end{tabular}
\end{table}

\subsection{Contrasts and Signal Exposure}
Positive-minus-negative win differences are $(29,26,24)$, or
$(48.33,43.33,40.00)$ percentage points; their equal-seed mean is
$+43.89$ points. Negative-minus-outcome differences are $(-37,-26,-35)$,
averaging $-54.44$ points. Positive-minus-outcome differences are $(-8,0,-11)$,
averaging $-10.56$ points. Under independent equiprobable nonzero seed signs,
two-sided exact sign-test values are $0.25$, $0.25$, and $0.50$, respectively.
The descriptive separation has three training-seed replicates; evaluation
games and optimization samples do not enlarge that replication count.

\begin{table}[!htbp]
\caption{Verified nonzero dense signals among samples used for optimization.
Denominators differ across conditions and are not independent experimental
replications.}
\label{tab:direction100-exposure}
\centering\small
\begin{tabular}{lrrr}
\toprule
Condition & Nonzero dense samples & Optimization samples & Fraction\\
\midrule
Outcome only & 0 & 406,029 & $0.00\%$\\
Positive oracle & 121,637 & 395,884 & $30.73\%$\\
Negative oracle & 300,588 & 680,933 & $44.14\%$\\
\bottomrule
\end{tabular}
\end{table}

The positive and negative providers differ only by sign, while visited
states and trajectory lengths differ after learning. Thus nonzero-sample
fractions need not match despite equal episode and update budgets. Sample
counts and cumulative dense returns reconcile with the update records.

The provider calls the historical \texttt{c4\_oracle.analyze}, whose defense
predicate checks whether an action occupies a pre-action threat column.
It predates the successor-state correction in
Appendix~\ref{app:reference-revalidation}. Consequently, ``positive oracle''
denotes the original sign of that historical rule, not a reward proven
tactically correct under the revised reference.

Evaluation uses fixed schedules and three opponents, with 20 games per
opponent and 10 per seat at each checkpoint. The random-opponent component
reuses first-update training RNG seeds: all 20 random-opponent evaluation
seeds per training seed fall inside that seed's first 200 training episodes.
This is schedule reuse; learned actions can still yield different paths.
The control measures sign sensitivity under that fixed schedule, rather
than performance on a newly sampled opponent schedule.

The result supports reward-sign sensitivity in this
control learner: the negative condition degrades in every seed relative to
both alternatives. Positive oracle shaping gives no observed gain over
outcome-only learning. The historical five-update result and this step-100
control are distinct batches; their difference alone does not isolate training
length. The control remains separate from the historical teacher-action objective
audit and from the completed clean Connect Four on-policy reference; it tests
reward-sign sensitivity in the traditional learner under its own fixed
schedule.

\subsection{Artifacts and Verification Scope}
The evidence directory is
\begin{quote}\small
\path{evidence/c4_reward_direction_20260908/}
\end{quote}
It contains the report (\path{REPORT.md}), aggregate results
(\path{aggregate_summary.json}), learning curves in CSV and JSONL
(\path{learning_curves.csv} and \path{learning_curves.jsonl}), and the hash
manifest (\path{artifact_hashes.json}). Under
\path{locusrl_environment_scaffold/}, the source files are
\begin{quote}\small
\path{locusrl/training/c4_reward_direction_experiment.py}\\
\path{locusrl/llm_rewards/c4_oracle_direction.py}
\end{quote}
Each of nine condition--seed runs contains six fixed checkpoint nodes,
100 training-metric records, complete optimization reward samples, and
360 evaluation records. The audit directly checks the provider sign and
terminal dense-reward rules. The submitter also reports ten current tests
and 19 targeted tests; those full suites were not rerun during the artifact
audit. The completed sign control does not require a further experiment
to validate its descriptive record.

\section{Leduc Normalized-Policy Training and Fresh-Schedule Evaluation}
\label{app:leduc-policy}

\subsection{Task, Learners, and Corrected Update Identity}
This extension uses the frozen simplified Leduc adapter: each player has a
hidden private card, a shared public card appears after the first betting
round, and re-raising is disabled. Prompts and Base features contain only
the acting player's information state. Terminal returns lie in $[-7,+7]$.
The reward is the environment outcome; no LLM reward cache or online judge
is used, so the reward-release decisions in Appendix~\ref{app:leduc-gate}
remain separate.

Base is a masked-action MLP with two 64-unit hidden layers and Adam learning
rate $10^{-3}$. Policy adapts \texttt{Qwen2.5-3B-Instruct} with 4-bit NF4
double quantization and rank-16 LoRA (scale 32), using AdamW at
$2\times10^{-4}$. Each arm has three training seeds, 200 episodes and
100 optimizer steps per seed, with two episodes per update. The training
opponent is \texttt{random\_v1}; the agent seat alternates with episode ID.
Policy scoring has a 512-token context limit. The two arms differ in
architecture and initialization, so this is a comparison of learning systems.

Let $\mathcal A(o)$ be the legal action set and $s_\theta(a,o)$ the model's
log probability of the completion \texttt{ACTION: a}, including completion
termination tokens but excluding prompt tokens. The actual policy is
\begin{equation}
 \pi_\theta(a\mid o,\mathcal A)
 =\frac{\exp s_\theta(a,o)}{\sum_{b\in\mathcal A(o)}\exp s_\theta(b,o)},
 \qquad
 \mathcal L(\theta)=-\frac{1}{M}\sum_{i=1}^{M}A_i
 \log\pi_\theta(a_i\mid o_i,\mathcal A_i).
 \label{eq:leduc-normalized-policy}
\end{equation}
Here $M$ counts agent decisions in the two-episode update batch. Each
decision uses its episode's terminal return minus the batch mean; when both
returns are identical, the implementation uses the terminal returns directly,
including zero for an all-zero batch. This specifies the implemented
batch-centered estimator rather than asserting an unbiased estimator of
expected episode return.

The first implementation sampled from $\pi_\theta$ but optimized
$-A_i s_\theta(a_i,o_i)$, omitting the normalization derivative. It also
enabled LoRA dropout during updates after disabling it for sampling.
The repaired implementation differentiates through every legal candidate
in Equation~\ref{eq:leduc-normalized-policy} and uses evaluation mode during
both sampling and gradient computation. Gradients remain enabled for the
update. A real-model smoke precedes three full Policy reruns; all three
record zero discrepancy between sampled and recomputed pre-update log
probabilities. Parameter-change and checkpoint-reload records are retained.
Base weights are reused from the original terminal-reward run.

\subsection{Frozen Evaluation and Replication Units}
The final evaluator samples from the same legal-action categorical policy
after checkpoint reload. Each seed--arm cell contains 60 games against
\texttt{random\_v1}, split evenly between seats. For training seed $s$, let
$e(s)$ be the evaluation seed in Table~\ref{tab:leduc-fresh-results}. Episode
$j\in\{0,\ldots,59\}$ uses environment seed $e(s)+j$, opponent seed
$e(s)+100003+j$, and agent seat $j\bmod2$. The Torch sampling seed is
$e(s)+700000+j$; the same schedule is used by both arms.

Training uses environment seed $s+i$ for $i\in\{0,\ldots,199\}$. An earlier
evaluation set $e(s)=s+10$, unintentionally reusing the training reset and
opponent RNG streams at episode IDs 10--69, with the same seats. That
fixed-schedule result is retained separately. The final schedule below was
frozen before the new Base and Policy evaluations; it has no overlap with
training, the earlier evaluation, or another training-seed block. This
checks RNG-stream reuse, not uniqueness of information states in a finite
game. The final comparison uses the same six checkpoints without retraining.

\begin{table}[!htbp]
\caption{Final fresh-schedule Leduc comparison. Each cell has 60 games,
30 per seat; W/D/L denotes positive/zero/negative terminal return.
The last column is the Policy-minus-Base difference in mean chips/game.}
\label{tab:leduc-fresh-results}
\centering\footnotesize
\setlength{\tabcolsep}{4pt}
\begin{tabular}{llrrrrr}
\toprule
Train seed & Eval seed & Base W/D/L & Policy W/D/L & Base sum & Policy sum & $\Delta$\\
\midrule
20260800 & 30260800 & 43/6/11 & 45/8/7 & 80 & 84 & $+0.0667$\\
20260801 & 30261800 & 37/6/17 & 19/17/24 & 18 & $-11$ & $-0.4833$\\
20260802 & 30262800 & 42/4/14 & 44/5/11 & 42 & 51 & $+0.1500$\\
\midrule
Total & --- & 122/16/42 & 108/30/42 & 140 & 124 & $-0.0889$\\
\bottomrule
\end{tabular}
\end{table}

Base and Policy average $140/180=0.7778$ and $124/180=0.6889$ chips/game.
Their win rates are $67.78\%$ and $60.00\%$, while both lose 42 games;
Policy has 30 draws compared with Base's 16. Two seed return contrasts are
positive and one negative. Three paired training seeds are the replication
blocks; 360 game records do not increase the number of independent training
replicates. Under independent equiprobable nonzero contrast signs, the
two-sided exact sign-test value is $p=1.0$. This small experiment establishes
neither a general improvement nor equivalence between the systems.

\subsection{Behavioral Diagnosis on Visited States}
\begin{table}[!htbp]
\caption{Policy action patterns on evaluated states. An opportunity is a
visited agent decision at which \texttt{bet\_raise} is legal. These counts
describe endogenous visited-state distributions, not additional replicates.}
\label{tab:leduc-action-patterns}
\centering\small
\begin{tabular}{lrrrr}
\toprule
Train seed & Agent decisions & Raise opportunities & Raises & Checks/calls\\
\midrule
20260800 & 99 & 68 & 68 & 31\\
20260801 & 155 & 89 & 0 & 155\\
20260802 & 98 & 68 & 68 & 30\\
\bottomrule
\end{tabular}
\end{table}

The first and third Policy checkpoints raise whenever that action is
available on the evaluated paths; the second always checks/calls. None folds
on these paths. These distinct patterns provide observable behavior alongside
the mixed return effects. They are not a complete characterization of all
reachable information states or an isolated causal explanation of returns.
Executed legality is $100\%$ by constrained candidate selection; free-text
generation and parsing are not part of this evaluation.

\subsection{Artifacts and Reproduction}
The normalized training and fresh evaluation are archived in
\begin{quote}\small
\path{outputs/leduc_terminal_two_arm_20260912_normalized/}\\
\path{outputs/leduc_terminal_two_arm_20260912_fresh_rng/}
\end{quote}
under \path{locusrl_environment_scaffold/}. Reported archive commits are
\texttt{ac5c9f2} for the normalized training and \texttt{5c59fb1} for the
fresh Policy evaluation. Base checkpoints remain under
\path{outputs/leduc_terminal_two_arm_20260911/formal/}.
The Qwen snapshot revision is
\begin{quote}\small
\texttt{aa8e72537993ba99e69dfaafa59ed015b17504d1}.
\end{quote}
The exact training and evaluation source hashes are:
\begin{quote}\small\ttfamily
eb257d00369bd4604ae8773f85bcf32584\\
e17a0b32e0154d39fd94c63b48cb4a\\[2pt]
c1a91cb9708803ece1906cf75112819c8d\\
0ab0866861be4ff80dd8fbabc4160c
\end{quote}

The artifact audit matches all 22 files in the fresh-evaluation checksum
manifest and confirms the Base records are byte-identical to the supplied
CPU evaluation. It replays all 360 games and 1,359 actions, verifies permitted
observations and payoff accounting, and matches checkpoint identities and
pairing keys. Logged Policy probabilities agree with score-based
recalculation within $6.03\times10^{-8}$. The preceding normalized-training
audit checked 600 training games and 300 updates. Seventeen CPU regression
tests were passed during the evaluator repair; the remote supplement
reports 11 regression tests. Real Qwen weight updates and reload are
supported by the run logs; the artifact audit does not independently rerun
GPU inference from the LFS adapters.

From the scaffold directory, the frozen final evaluation can be run using
the local model snapshot and the archived checkpoints:
\begin{quote}\footnotesize\ttfamily
PYTHONPATH=. python scripts/leduc\_two\_arm\_eval\_20260912.py\ \textbackslash\\
\quad --model /path/to/Qwen2.5-3B-Instruct\ \textbackslash\\
\quad --base-root outputs/leduc\_terminal\_two\_arm\_20260911/formal\ \textbackslash\\
\quad --policy-root /path/to/normalized/formal\ \textbackslash\\
\quad --output-dir outputs/leduc\_fresh\_reproduction --arm all
\end{quote}
Here \path{/path/to/normalized/formal} denotes the normalized Policy root
listed above, and the output directory must be new. The standalone
\path{verify_leduc_fresh_results.py} checks the supplied fresh-evaluation
records without loading Qwen or calling an API:
\begin{quote}\footnotesize\ttfamily
python verify\_leduc\_fresh\_results.py /path/to/scaffold\ \textbackslash\\
\quad --output combined\_summary.json
\end{quote}
The combined summary identifies all six source episode files and their
hashes, per-seed results, seed-level contrasts, and the completed two-arm
scope. The two original single-arm summaries are preserved separately.

\end{document}